\documentclass{article}

\usepackage{arxiv_conference,times}
\usepackage{amsmath,amssymb}
\usepackage{booktabs}
\usepackage{tabularx}
\usepackage{graphicx}
\usepackage{xcolor}
\usepackage{microtype}
\usepackage{float}
\usepackage{wrapfig}
\usepackage{needspace}
\usepackage{url}
\usepackage[hidelinks]{hyperref}

\newcommand{\TestEpisodes}{5,337}
\newcommand{\TestSubjects}{3,827}

\newcommand{\TrainEpisodes}{42,669}
\newcommand{\TrainSubjects}{30,835}

\newcommand{\ValEpisodes}{5,323}
\newcommand{\ValSubjects}{3,848}

\newcommand{\method}{PatientWM}

\newcommand{\latent}{\mathbf{z}}
\newcommand{\action}{\mathbf{a}}

\title{PatientWM: Towards Generalizable Clinical World Models via Unified Patient States}
\author{
  Ziwen Kan \quad Jie Peng$^{2}$ \quad Bo Yan$^{6}$ \quad Davis Hobley$^{4}$ \\
  Zheng Zhang$^{5}$ \quad Yu Tian$^{6}$ \quad Tianlong Chen$^{7}$ \quad Song Wang$^{6}$ \\
  $^{2}$University of Science and Technology of China \\
  $^{4}$Princeton University \\
  $^{5}$Nanyang Technological University \\
  $^{6}$University of Central Florida \\
  $^{7}$University of North Carolina at Chapel Hill
}

\arxivfinalcopy

\begin{document}
\maketitle

\begin{abstract}
Clinical decisions depend on anticipating how a patient's condition will evolve as treatment is adjusted and new evidence becomes available. Existing medical world models capture patient-state evolution, but often define the evolving state around a particular modality or task, limiting reuse as inputs or prediction targets change. These models also retain history through observation sequences or recurrent or latent summaries rather than exposing preceding patient states and clinical actions as an ordered transition context. This distinction matters because similar current measurements may follow different recent changes, treatment exposures, and observed responses. To tackle these limitations, we propose PatientWM, a clinical world-model framework built around a unified patient state and sequential latent history. Availability-aware encoders integrate observations available at each decision time into the patient state, while typed decoders produce task-specific clinical readouts without defining the state around any single target. PatientWM further conditions each next-state prediction on the current state and supplied action together with a bounded, ordered history of preceding latent patient states and clinical actions, preserving recent treatment and response context.
Experiments across MIMIC-III, MIMIC-IV, and three external clinical datasets show superior performance in clinical prediction, multi-step forecasting, and action-planning evaluation. Transfer experiments further demonstrate the generalizability of PatientWM's patient state. Together, these results establish its ability to forecast clinical evolution and support planning.
\end{abstract}

\section{Introduction}

\begin{figure*}[!t]
  \centering
  \includegraphics[width=0.98\textwidth]{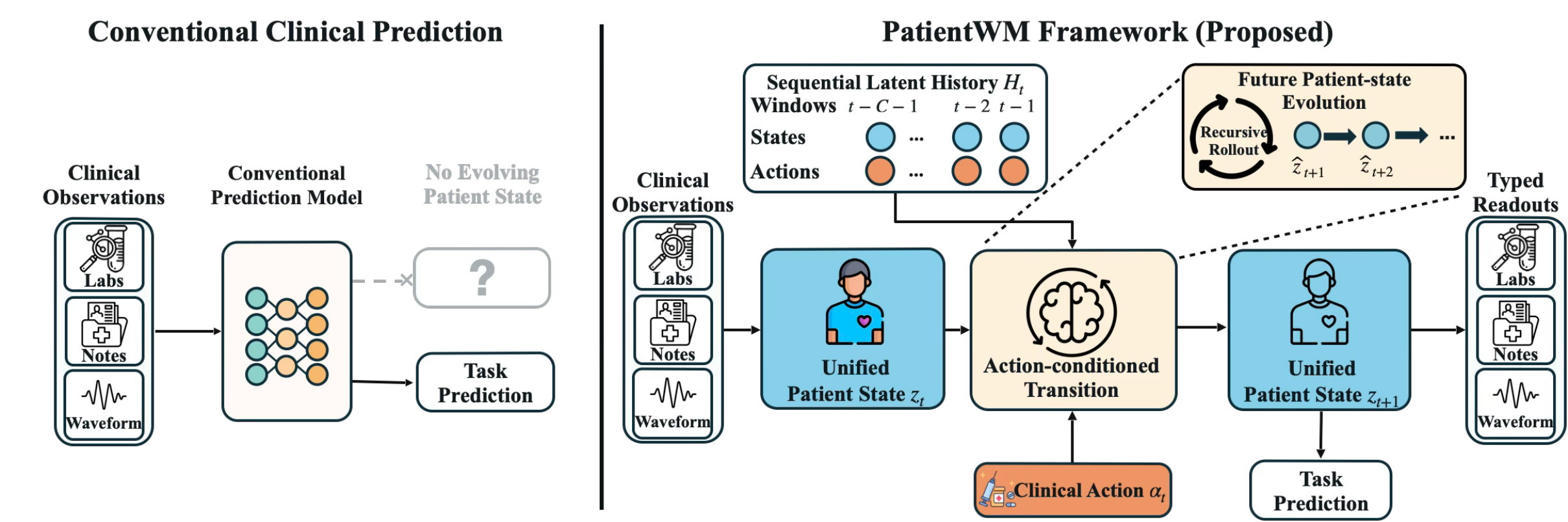}
  \caption{\textbf{Conceptual Overview of PatientWM.} Conventional clinical prediction focuses on learning task-specific predictions from heterogeneous observations. PatientWM instead learns a unified patient state from available observations and models how this state evolves under clinical actions and recent history. The learned state supports multiple clinical readouts and multi-step forecasting.}
  \label{fig:teaser}
\end{figure*}

Clinical treatment often unfolds as a sequential decision process, in which clinicians repeatedly select or adjust treatments based on a patient's current health status, prior treatment history, and observed responses over time~\citep{yu2021reinforcement}. Each decision depends not only on the observations available at that moment, but also on how the patient's condition has evolved under preceding treatments~\citep{shickel2017deep}. Conventional clinical AI systems commonly map longitudinal observations directly to predefined, task-specific outcomes, such as diagnoses or risk estimates~\citep{choi2016retain,rajkomar2018scalable}.
As illustrated in Figure~\ref{fig:teaser}, such outputs answer a particular prediction question, but do not by themselves define a shared patient state that can be updated after treatment and carried forward to model subsequent clinical evolution.

Modeling this evolution requires representing the patient's condition at the current decision time and predicting how that state changes as treatments are applied. World models are well suited to this setting because they represent a system's current state and model its action-conditioned dynamics to predict future states. These models have been applied in autonomous driving, robotics, video generation, and social simulation to forecast future states and support planning or evaluation~\citep{ha2018worldmodels,ding2025worldmodelsurvey}. In clinical settings, recent studies have begun to utilize this framework to model patient trajectories for specialized clinical modalities~\citep{liu2026medical,chen2026medicalworldreview,mu2026ehrworld,xu2025meddreamer,ding2026clarity,wang2026chronomedicalworld}.

Despite this progress, two challenges remain for the formulation considered here. (1) \textbf{Limited-Scope State}: Existing medical world models typically construct patient states for a fixed set of clinical inputs and prediction tasks. This makes the learned state harder to reuse when additional modalities or new prediction targets are introduced~\citep{si2021deep,hansen2024multimodal,dou2025survey,liu2026medical}. (2) \textbf{Implicit History}: Existing methods generally retain history through raw observation sequences or compress it into recurrent or latent context, without making the ordered sequence of preceding patient states and clinical actions separately available to each transition~\citep{wu2026sepsisagent,xu2025meddreamer,ding2026clarity,wang2026chronomedicalworld}. Consequently, it can be challenging for the transition model to learn how similar current states reached through different treatment-response trajectories may evolve differently.

To address these limitations, we propose PatientWM, a clinical world model built around a \textbf{unified patient state} that represents the patient's current condition, supports prediction of its future evolution, and serves as a shared basis for downstream clinical predictions. At each decision time, modality-specific encoders represent the available clinical observations, which are fused into the unified patient state, while typed decoders produce task-specific outputs on demand. PatientWM further represents recent history as \textbf{sequential latent history}: an ordered sequence of latent patient-state and clinical-action pairs. Each next-state prediction then conditions on this sequence together with the current state and supplied action. This explicit trajectory context helps the transition model distinguish similar current states reached through different treatment-response histories, which may imply different future clinical evolution.
As illustrated in Figure~\ref{fig:teaser}, PatientWM models patient evolution directly in the unified state space, with sequential latent history and the supplied action conditioning each predicted state transition.

Our main contributions are summarized below.
\begin{itemize}
  \item We propose PatientWM, a clinical world-model framework built around a unified patient state. It learns this state from available clinical observations, models patient evolution in the shared state space, and uses typed decoders to produce task-specific outputs on demand.
  \item We represent patient history as an ordered sequence of latent patient-state and clinical-action pairs. Each transition conditions on this sequence together with the current state and supplied action, making trajectory context an explicit part of patient-state evolution.
  \item Experiments across MIMIC-III, MIMIC-IV, and three external clinical datasets show superior performance in clinical prediction, multi-step forecasting, and action-planning evaluation. Transfer experiments further demonstrate the generalizability of PatientWM's patient state. These results demonstrate its ability to forecast clinical evolution and support planning.
\end{itemize}

\section{Related Work}

\paragraph{Clinical Prediction.} Clinical prediction models use longitudinal EHRs and physiological signals~\citep{shickel2017deep,harutyunyan2019multitask,faust2018physiological} to predict diagnoses, deterioration, clinical events, and outcomes across diverse clinical settings~\citep{choi2016retain,tomasev2019aki,rajkomar2018scalable,steinberg2023motor,hannun2019arrhythmia,attia2019screening}. These systems capture temporal and physiological patterns, but are typically organized around predefined diagnostic or prognostic targets. Such outputs do not by themselves represent how a patient's condition evolves between observations and treatment updates, which is central to sequential clinical decision support~\citep{liu2026medical,chen2026medicalworldreview}.

\paragraph{World Models.} World models include latent state-space models such as World Models and Dreamer for imagined control~\citep{ha2018worldmodels,hafner2019dreamer,hafner2025dreamerv3}, extensions such as DayDreamer for physical robot learning~\citep{wu2023daydreamer}, and generative models such as Drive-WM and Genie for controllable environments~\citep{wang2024drivewm,bruce2024genie}. Across these settings, the common principle is to connect an internal representation of the environment with action-conditioned future prediction, allowing possible trajectories to be simulated for planning, policy learning, or evaluation~\citep{ding2025worldmodelsurvey}. This principle has recently been adapted to medicine by modeling patient states and their evolution under clinical interventions~\citep{liu2026medical}.

\paragraph{Medical World Models.} Medical world models and related trajectory models differ in what they treat as the evolving patient state. Some models define the evolving state directly in observable clinical space. EHRWorld and EHR2Path~\citep{mu2026ehrworld,pellegrini2026ehr2path} operate in record or pathway space, generating subsequent chart events or pathway updates from the preceding trajectory. SepsisAgent~\citep{wu2026sepsisagent} predicts physiological evolution in structured observation space and remains tied to predefined clinical variables. Latent-state approaches include medDreamer~\citep{xu2025meddreamer}, which supports action-conditioned rollouts for offline reinforcement learning, and ACWM~\citep{fernandes2026acwm}, whose action labels denote physiological events rather than clinical treatments. Other medical world models target specialized imaging, postoperative, disease-specific, and physiological-signal settings~\citep{ding2026clarity,yang2025mewm,chung2026interventionaware,wang2026chronomedicalworld,luo2026normwear2}. PatientWM instead integrates heterogeneous observations into a unified patient state with typed readouts and conditions each transition on an ordered history of latent patient states and supplied clinical actions.

\section{Method}
\label{sec:method}

\subsection{Problem Formulation}
PatientWM models clinical evolution at discrete time steps. Let $O_t$ denote
the heterogeneous clinical observations available at step $t$, together with
their availability indicators. Because these records are shaped by measurement
and documentation processes, $O_t$ provides evidence about the patient but is
not the patient state itself.
PatientWM infers a learned predictive patient state $\latent_t$ from $O_t$.
This state supports prediction of subsequent clinical evolution; it is neither
an observed modality nor a claim about the patient's true physiological state.
Let $\action_t$ denote the clinical action supplied for the transition from
step $t$ to $t+1$, and let $H_t$ denote the recent state--action history. The factual transition target is
\begin{equation}
  p^*\!\left(
    \latent_{t+1}
    \mid \latent_t,\action_t,H_t
  \right),
  \label{eq:latent_transition_target}
\end{equation}
where $p^*$ denotes the conditional distribution induced by the observed
clinical process, not an interventional treatment-response law. The modeled
object is the next predictive state in the shared latent space, while typed
clinical observations ground and evaluate this state. The formulation specifies
a conditional target, whereas the implemented transition produces a
deterministic prediction toward a stopped-gradient momentum target.

\begin{figure*}[!t]
  \centering
  \includegraphics[width=0.97\textwidth]{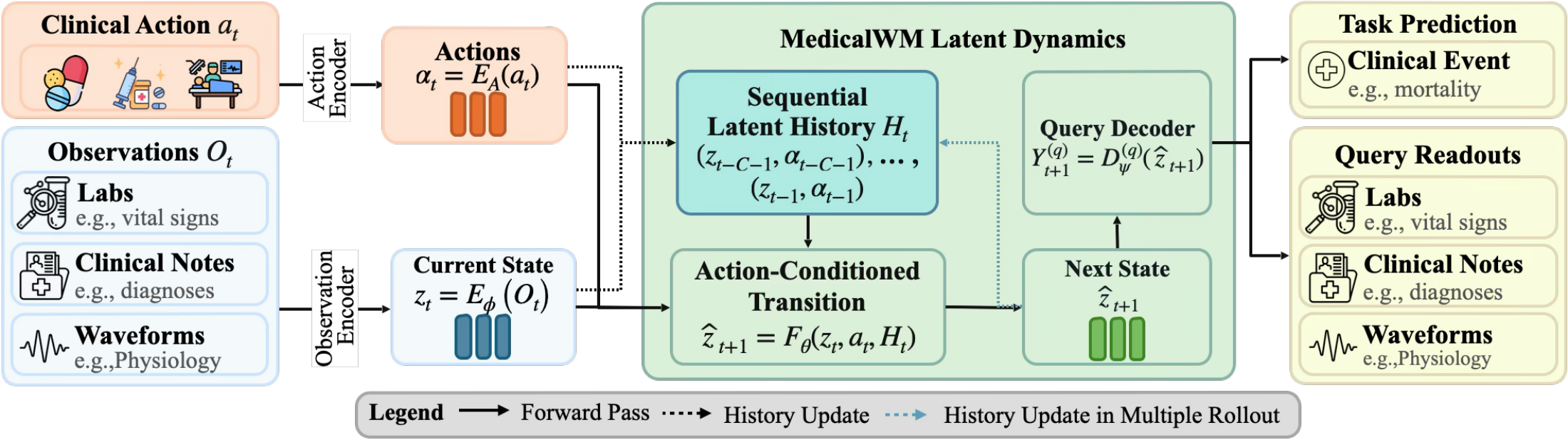}
  \caption{\method{} encodes available observations and the supplied action into a unified patient state $z_t$ and action representation $\alpha_t$. Conditioned on $z_t$, $\alpha_t$, and sequential latent history $H_t$, the transition predicts $\widehat z_{t+1}$. The query decoder produces clinical event and outcome predictions and query-specific readouts, including laboratory values. Black arrows show the training-time forward pass with observed history; blue arrows show recursive multi-step rollout, where predicted states update the history for subsequent transitions.}
  \label{fig:overview}
\end{figure*}

\subsection{State Representation}

\paragraph{Availability-Aware Observation Encoding.}
Let $\mathcal M_t$ be the modalities available at step $t$, with $m$ indexing
one modality, $o_t^m$ its observation, and $h_t^m$ its encoded representation.
The modality-specific encoders $E_m$ are fused by $E_O$ through
$f_{\mathrm{fusion}}$ to obtain the predictive patient state:
\begin{equation}
  h_t^m=E_m(o_t^m),
  \qquad
  \latent_t=E_O(O_t)
  =f_{\mathrm{fusion}}\left(
    \{h_t^m:m\in\mathcal M_t\}\right).
  \label{eq:modality_fusion_encoder}
\end{equation}
The fusion uses only modalities available at that step. The resulting
$\latent_t$ is the unified predictive patient state used by the transition and
typed readouts.

\paragraph{Typed Clinical Readouts.}
For target types $q\in\mathcal Q$, let $Y_{t+1}^{(q)}$ and
$\widehat Y_{t+1}^{(q)}$ denote the observed and predicted targets. A
type-specific decoder $D_\psi^{(q)}$ maps the predicted state to the
corresponding target:
\begin{equation}
  \widehat Y_{t+1}^{(q)}
  =D_\psi^{(q)}(\widehat\latent_{t+1}).
  \label{eq:typed_observation_readout}
\end{equation}
These decoders ground the shared patient state in observable clinical
quantities without defining that state around any single prediction target.
Here $q$ indexes clinical event, outcome, and continuous laboratory targets,
each with its corresponding readout and loss. Sharing $\latent_t$ makes it a
common predictive substrate whose dynamics support multiple clinical queries,
rather than an endpoint-specific representation.
Equation~\ref{eq:typed_observation_readout} describes the shared readout
interface. Continuous heads may additionally use the transition representation
or an output-specific causal prior, which remain decoder-side inputs rather
than additional shared patient states.

\paragraph{Persistence-Aware Continuous Readouts.}
For a continuous target, $r_{t+1}$ and $\widehat r_{t+1}$ denote the observed
and predicted values. PatientWM predicts a residual $\Delta_t^r$ around the
persistence prior $r_t^{\mathrm{prior}}$, controlled by a learned gate $g_t^r$:
\begin{equation}
  \widehat r_{t+1}
  =
  r_t^{\mathrm{prior}}
  +
  g_t^r\odot\Delta_t^r.
  \label{eq:persistence_aware_readout}
\end{equation}
Here $\odot$ denotes element-wise multiplication. The gate controls how much
of the predicted residual is applied to the prior.
The waveform prior uses the most recent quality-qualified embedding with a
learned fallback when none exists. The laboratory prior uses the last observed
normalized value and train-mean zero for a never-observed item. These priors
remain within their typed readouts and do not alter the recursively reused
shared-state slot.

\subsection{Action-Conditioned Latent Dynamics}

\paragraph{Sequential Latent History.}
The supplied clinical action is encoded separately, and PatientWM retains the
preceding $C-1$ latent-state and action representations:
\begin{equation}
  \boldsymbol{\alpha}_t=E_A(\action_t),
  \qquad
  H_t=
  \bigl(
    (\latent_s,\boldsymbol{\alpha}_s)
  \bigr)_{s=t-C+1}^{t-1}.
  \label{eq:latent_history}
\end{equation}
Together with $(\latent_t,\boldsymbol{\alpha}_t)$, this provides each transition
with an ordered context of at most $C$ steps. Encoding the action separately
keeps treatment-side information explicit without assuming independence from
the observation-derived state. The ordered history distinguishes transitions
that reach similar current states through different preceding state--action
paths, rather than requiring $\latent_t$ to compress the complete trajectory.

\paragraph{Residual Transition.}
PatientWM predicts the next patient state from the current state, supplied
action, and sequential latent history:
\begin{equation}
  \widehat\latent_{t+1}
  =
  F_\theta(\latent_t,\boldsymbol{\alpha}_t,H_t)
  =
  \latent_t+
  \Delta_\theta(\latent_t,\boldsymbol{\alpha}_t,H_t).
  \label{eq:action_conditioned_transition}
\end{equation}
The residual form parameterizes state changes relative to the current
predictive patient state, with $\Delta_\theta$ representing the complete
action- and history-conditioned update. The formulation is architecture
agnostic; our implementation uses a causally masked sequence model and learned
gate to retain supported information while applying predicted changes. Further
details are provided in the appendix.

\paragraph{Training Objective.}
To provide a stable target patient state for learning the latent transition, we maintain a momentum-updated copy $\bar E_O$ of the observation encoder, which encodes the next-step observations as follows:
\begin{equation}
  \latent_{t+1}^{\mathrm{tar}}
  =
  \operatorname{stopgrad}\!\left(
    \bar E_O(O_{t+1})
  \right).
  \label{eq:latent_training_target}
\end{equation}
The momentum encoder provides a slowly evolving shared-state target, while the
typed objectives ground it in clinically observable quantities. The one-step
objective is
\begin{equation}
  \mathcal L_{\mathrm{direct}}
  =
  \mathcal L_z(\widehat\latent_{t+1},
               \latent_{t+1}^{\mathrm{tar}})
  +\sum_{q\in\mathcal Q}
    m_{t+1}^{(q)}
    \lambda_q
    \mathcal L_q\!\left(
      \widehat Y_{t+1}^{(q)},
      Y_{t+1}^{(q)}\right).
  \label{eq:direct_latent_readout_loss}
\end{equation}
Here $m_{t+1}^{(q)}$ indicates whether target $q$ is observed and included in
the loss, and $\lambda_q$ weights its type-specific objective. The latent loss
aligns the predicted and target patient states, while the typed losses ground
the transition in observable clinical quantities.

\subsection{Factual Rollout}

\paragraph{Recursive State Evolution.}
To model evolution beyond the next step, PatientWM recursively reuses its
predicted patient states. Starting from an observed state
$\widehat\latent_{t_0}=\latent_{t_0}$, it computes
\begin{equation}
  \widehat\latent_{t_0+k+1}
  =
  F_\theta\!\left(
    \widehat\latent_{t_0+k},
    \boldsymbol{\alpha}_{t_0+k},
    \widehat H_{t_0+k}
  \right),
  \qquad k=0,\ldots,K-1 .
  \label{eq:factual_latent_rollout}
\end{equation}
where $\widehat H_{t_0+k}$ contains the retained observed or previously
predicted state--action pairs. The history is truncated according to
Eq.~\ref{eq:latent_history}, and the action at each step remains externally
supplied. After initialization, each predicted state is appended to the
retained history and becomes the state input to the next transition. Future
observations are not re-encoded during rollout, distinguishing recursive state
evolution from repeated one-step prediction with observed history.

\paragraph{Recursive Training.}
Under the recursive training configuration, the direct objective predicts the next state from observed history, while an additional rollout term feeds generated states into subsequent transitions to supervise multi-step prediction:
\begin{equation}
  \mathcal L_{\mathrm{rollout}}
  =
  \mathcal L_{\mathrm{direct}}
  +
  \lambda_r(t)
  \sum_{k=1}^{H}
  \mathcal L_{\mathrm{direct}}^{(t+k\mid t)},
  \label{eq:recursive_rollout_loss}
\end{equation}
where $H$ is the training rollout horizon and $\lambda_r(t)$ controls the
recursive loss contribution. Decoder-side priors are carried across rollout
steps but do not constitute complete write-back of generated clinical
observations; further details are provided in the appendix.

\section{Experiments}

\subsection{Experimental Setup}
\label{sec:experimental_setup}

\paragraph{Datasets and Transition Unit.}
We evaluate PatientWM on MIMIC-III and MIMIC-IV clinical trajectories,
including time-aligned waveform inputs where available
\citep{johnson2016mimic,moody2020mimicwaveform,johnson2023mimic}.
Both datasets are organized into contiguous six-hour event threads:
observations and source-available treatment records at step $t$ form the model
input, while the following window provides factual state and readout targets.
Dataset construction, cohort statistics, action definitions, and information
boundaries are detailed in Appendix~\ref{app:datasets}.

\paragraph{Comparisons and Evaluation Roles.}
Our experiments evaluate factual prediction through typed clinical readouts,
dependence on the proposed model components, recursive reuse of generated
patient states, action-planning evaluation over bounded candidates, and
generalization through frozen-state reuse, label-efficient cross-dataset
adaptation, and external cohorts. We compare SepsisAgent, medDreamer, ACWM, and
SMB-Structure through the same clinical prediction interface. Dataset-facing
input layers allow each method to consume the available laboratory and
clinical-fact inputs together with embeddings from the same frozen waveform
encoder, while preserving its defining dynamics or representation core. All
methods use the same visible inputs, action timing where applicable, target
definitions, eligible support, and evaluator. The
comparisons therefore assess the retained method structures under matched
multimodal information.
Implementation details are provided in Appendix~\ref{app:comparison_details}.

\paragraph{Metrics.}
We report AUPRC~\citep{saito2015precision},
AUROC~\citep{davis2006relationship}, and Brier
score~\citep{brier1950verification} for clinical events and outcomes,
normalized MAE (nMAE) for observed laboratory values, and coverage and top-$k$
agreement for the action-planning evaluation.
Metrics remain target-specific rather than being combined into a heterogeneous
aggregate. Full protocols and metric definitions are provided in
Appendix~\ref{app:evaluation_protocol}.

\subsection{Factual Prediction And Clinical Readout}

\paragraph{Clinical Targets.}
We evaluate clinical prediction in three panels: dataset-specific MIMIC-III
Core-7 outcomes and a shared Common Event-7 interface on MIMIC-III and
MIMIC-IV (Table~\ref{tab:targetbenchmark}). The shared panel combines six acute
next-window events present in both datasets with dynamic ICU mortality at its
native 24-hour horizon. PatientWM achieves the strongest overall aggregate
performance across all three panels. Figure~\ref{fig:core7_event7_radar} and
the corresponding AUPRC and Brier analyses in
Appendix~\ref{app:outcome_configurations} further characterize its performance
across individual clinical events and outcomes. Together, these results
demonstrate the effectiveness of PatientWM for clinical prediction across
heterogeneous targets and datasets.

\begin{table*}[!t]
  \centering
  \small
  \setlength{\tabcolsep}{3pt}
  \caption{{Clinical event and outcome prediction on the MIMIC-III and MIMIC-IV test sets. The three panels report MIMIC-III Core-7 outcomes and Common Event-7 readouts on both datasets. Bold indicates the best result in each column.}}
  \label{tab:targetbenchmark}
  \vspace{.5em}
  \resizebox{0.99\textwidth}{!}{\begin{tabular}{l *{9}{r}}
\toprule
& \multicolumn{3}{c}{\textbf{MIMIC-III: Core-7}} & \multicolumn{3}{c}{\textbf{MIMIC-III: Common Event-7}} & \multicolumn{3}{c}{\textbf{MIMIC-IV: Common Event-7}} \\
\cmidrule(lr){2-4} \cmidrule(lr){5-7} \cmidrule(lr){8-10}
Method & AUPRC$\uparrow$ & AUROC$\uparrow$ & Brier$\downarrow$ & AUPRC$\uparrow$ & AUROC$\uparrow$ & Brier$\downarrow$ & AUPRC$\uparrow$ & AUROC$\uparrow$ & Brier$\downarrow$ \\
\midrule
medDreamer       & 0.0579          & 0.5130          & 0.2192          & 0.0278          & 0.5049          & 0.2631          & 0.0341          & 0.5050          & 0.2547 \\
SMB-Structure    & 0.0638          & 0.5412          & 0.2165          & 0.0522          & 0.6000          & 0.2278          & 0.0604          & 0.6375          & 0.2572 \\
SepsisAgent      & 0.0772          & 0.6191          & 0.2181          & 0.0517          & 0.6631          & 0.2457          & 0.1742          & 0.7971          & 0.1875 \\
ACWM             & 0.0762          & 0.5945          & 0.2235          & 0.0444          & 0.5879          & 0.2444          & 0.1768          & 0.7959          & 0.1846 \\
PatientWM (Ours) & \textbf{0.2227} & \textbf{0.7974} & \textbf{0.0761} & \textbf{0.0598} & \textbf{0.6837} & \textbf{0.1964} & \textbf{0.1784} & \textbf{0.7988} & \textbf{0.1704} \\
\bottomrule
\end{tabular}}
\end{table*}

\begin{figure*}[!t]
  \centering
  \includegraphics[width=0.99\textwidth]{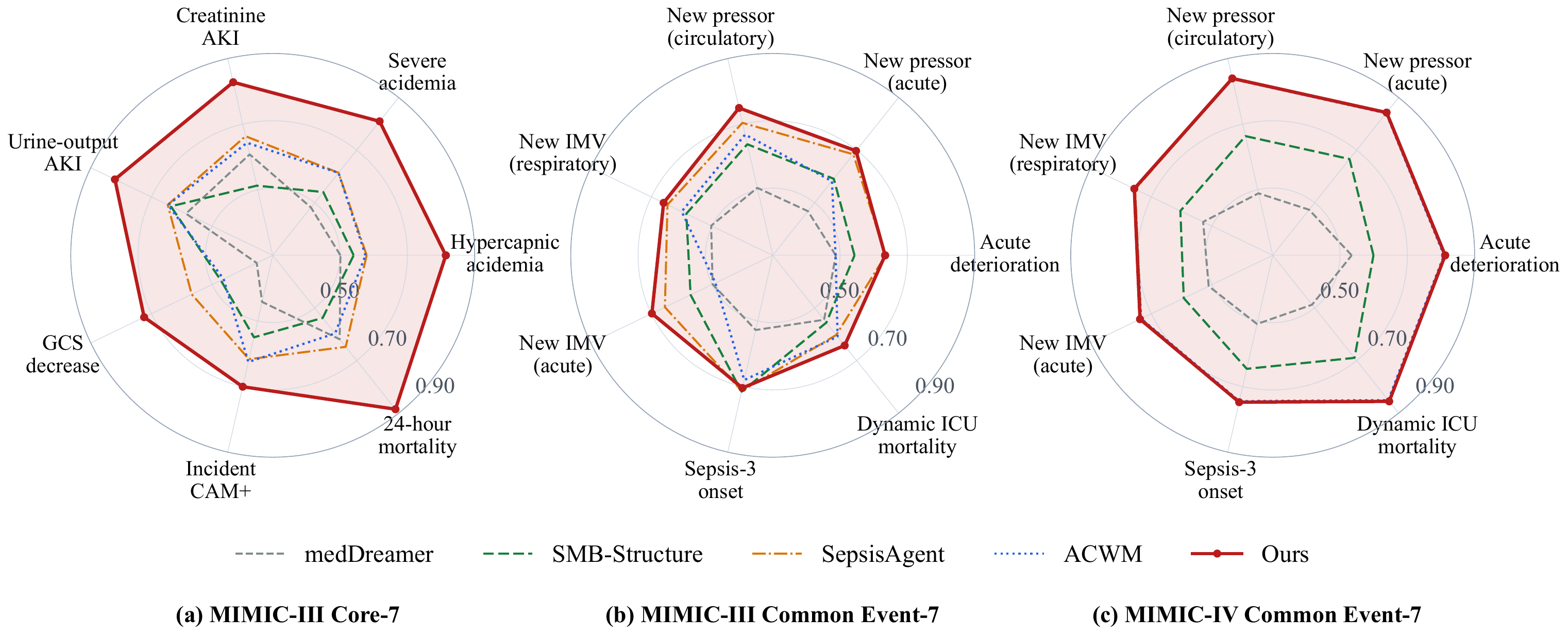}
  \caption{Per-target AUROC profiles for (a) MIMIC-III Core-7
outcomes, (b) MIMIC-III Common Event-7 readouts, and (c) MIMIC-IV Common
Event-7 readouts. Corresponding per-target AUPRC and Brier results are reported
in Appendix~\ref{app:outcome_configurations}.}
  \label{fig:core7_event7_radar}
\end{figure*}

\paragraph{Laboratory Values.}
PatientWM outperforms all baselines in laboratory-value prediction at every
evaluated horizon on both MIMIC-III and MIMIC-IV, as shown in Table~\ref{tab:mimiciv_matched_test}. These consistent gains from 6 to 96
hours demonstrate that PatientWM more accurately reconstructs future observed
laboratory values from the evolving patient state.

\begin{table*}[!t]
  \centering
  \small
  \caption{Laboratory-value prediction on the MIMIC-III and MIMIC-IV test cohorts across 6--96-hour horizons. Bold marks the best result within each panel.}
  \label{tab:mimiciv_matched_test}
  \vspace{.5em}
  \resizebox{0.86\textwidth}{!}{%
\begin{tabular}{lrrrrr}
\toprule
\multicolumn{6}{l}{\textbf{MIMIC-III Observed Laboratory Values}} \\
Method & 6 h nMAE $\downarrow$ & 24 h nMAE $\downarrow$ & 48 h nMAE $\downarrow$ & 96 h nMAE $\downarrow$ & Mean nMAE $\downarrow$ \\
\midrule
medDreamer  & 0.4296 & 0.4791 & 0.5387 & 0.6079 & 0.5138 \\
SMB-Structure & 0.6257 & 0.6236 & 0.6300 & 0.6286 & 0.6270 \\
% GapShrink & 0.3474 & 0.4170 & 0.4749 & 0.5303 & 0.4424 \\
% Last-value carry-forward & 0.3568 & 0.4495 & 0.5224 & 0.6011 & 0.4824 \\
SepsisAgent & 0.4295 & 0.4792 & 0.5388 & 0.6086 & 0.5140 \\
ACWM & 0.4296 & 0.4789 & 0.5386 & 0.6081 & 0.5138 \\
PatientWM (Ours) & \textbf{0.3248} & \textbf{0.3888} & \textbf{0.4430} & \textbf{0.4966} & \textbf{0.4133} \\

\midrule
\multicolumn{6}{l}{\textbf{MIMIC-IV Observed Laboratory Values}} \\
Method & 6 h nMAE $\downarrow$ & 24 h nMAE $\downarrow$ & 48 h nMAE $\downarrow$ & 96 h nMAE $\downarrow$ & Mean nMAE $\downarrow$ \\
\midrule
medDreamer & 0.4243 & 0.4527 & 0.5093 & 0.5728 & 0.4898 \\
SMB-Structure & 0.5879 & 0.5862 & 0.5926 & 0.6022 & 0.5922 \\
SepsisAgent & 0.4191 & 0.4520 & 0.5085 & 0.5690 & 0.4871 \\
ACWM & 0.4240 & 0.4519 & 0.5084 & 0.5729 & 0.4893 \\
PatientWM (Ours) & \textbf{0.4093} & \textbf{0.4161} & \textbf{0.4556} & \textbf{0.4958} & \textbf{0.4442} \\
\bottomrule
\end{tabular}}

\end{table*}

\subsection{Ablations and Mechanism Analysis}

\begin{wraptable}{r}{0.40\columnwidth}
  \vspace{-1.8\baselineskip}
  \centering
  \scriptsize
  \setlength{\tabcolsep}{2.2pt}
  \renewcommand{\arraystretch}{0.94}
  \caption{MIMIC-III Core-7 component and modality ablations.}
  \label{tab:t004_module_ablation}
  \vspace{-0.2em}
  \resizebox{\linewidth}{!}{\begin{tabular}{@{}lrrr@{}}
\toprule
\multicolumn{4}{@{}l}{\textit{MIMIC-III Core-7}} \\
Method & AUPRC $\uparrow$ & AUROC $\uparrow$ & Brier $\downarrow$ \\
\midrule
PatientWM (Ours) & \textbf{0.2227} & \textbf{0.7974} & 
\textbf{0.0761} \\
\midrule
Without action & 0.0861 & 0.6067 & 0.2024 \\
Without history & 0.0840  & 0.5507 & 0.2523  \\
\midrule
Laboratory only & 0.0800 & 0.5874 & 0.2171 \\
Clinical facts only & 0.0777 & 0.5888 & 0.2137 \\
Waveform only & 0.0772 & 0.5853 & 0.2198 \\
\bottomrule
\end{tabular}
}
  \vspace{-0.8\baselineskip}
\end{wraptable}

We conduct an ablation study to evaluate the effectiveness of PatientWM's key
transition components and modalities. Table~\ref{tab:t004_module_ablation}
compares five ablations on the MIMIC-III Core-7 outcomes. The first two remove
action conditioning or ordered history. PatientWM outperforms both, supporting
the contribution of these mechanisms to learning clinically informative
patient-state dynamics. The larger degradation without ordered history further
shows the importance of preserving recent treatment-response trajectories. The
remaining three variants restrict state construction to laboratory values,
clinical facts, or waveforms alone. Their consistently lower performance shows
that no single modality is sufficient and supports combining complementary
clinical modalities in the unified patient state.

\begin{figure*}[!t]
  \centering
  \includegraphics[width=0.92\textwidth]{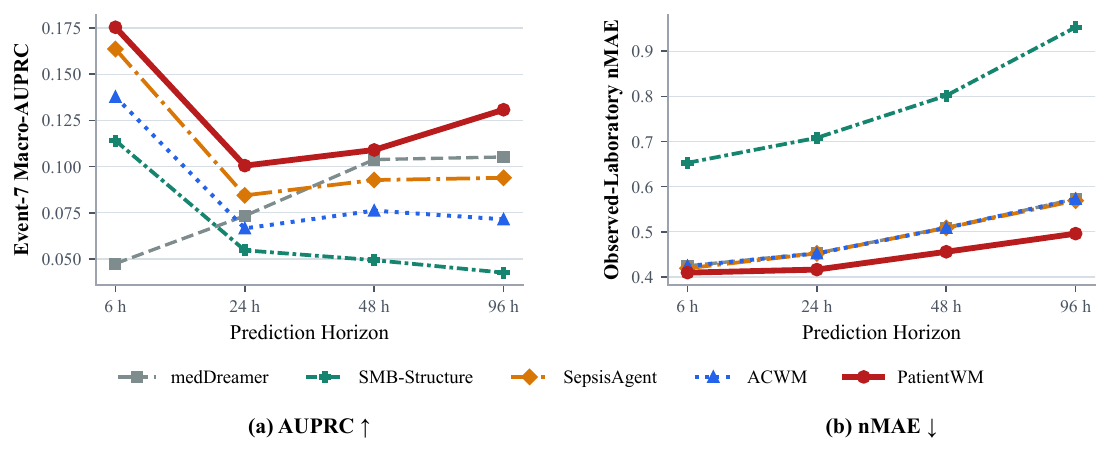}
  \caption{MIMIC-IV multi-step rollout comparison. Panel (a) reports recursive Event-7 macro-AUPRC, and panel (b) reports observed-laboratory nMAE from the typed laboratory decoder. Detailed results are reported in Appendix~\ref{app:rollout_details}.}
  \label{fig:mimiciv_strict_rollout}
\end{figure*}

\subsection{Multi-Step Patient-State Rollout}

Multi-step prediction requires models to propagate their own predicted states
across successive transitions. From the same observed initial states, we
recursively roll out PatientWM and baselines using recorded actions, while
reserving future clinical events and laboratory values solely for evaluation.
Although laboratory error increases and event ranking becomes more variable
with horizon, PatientWM achieves the highest Event-7 macro-AUPRC and the lowest
observed-laboratory nMAE at every horizon. These results support more
accurate continued forecasting under recorded actions. Figure~\ref{fig:mimiciv_strict_rollout}
summarizes the trend, with exact values and support audits reported in
Appendix~\ref{app:rollout_details}.

\Needspace{0.34\textheight}
\subsection{Action-Planning Evaluation}

\begin{wraptable}{r}{0.50\columnwidth}
  \vspace{-1.8\baselineskip}
  \centering
  \caption{Recorded-action agreement in the action-planning evaluation.}
  \label{tab:prediction_only_action_scoring}
  \vspace{.5em}
  \resizebox{\linewidth}{!}{\begin{tabular}{lrr}
\toprule
Method & Top-1 covered & Top-2 covered \\
\midrule
medDreamer & 0.4713 & 0.7997 \\
SMB-Structure & 0.3616 & 0.7925 \\
SepsisAgent & 0.6920 & 0.9081 \\
ACWM & 0.3576 & 0.7833 \\
PatientWM (Ours) & \textbf{0.7477} & \textbf{0.9126} \\
\bottomrule
\end{tabular}
}
  \vspace{-0.8\baselineskip}
\end{wraptable}

To compare action-planning capability, we evaluate whether predicted future
consequences can support ranking a bounded set of candidate clinical actions.
For every candidate, each method predicts the subsequent
clinical-event and laboratory burdens without access to the observed future;
the predicted burdens are then combined by the same fixed scoring rule. For
decisions in which the clinician's recorded action is represented in the
candidate set, covered top-$k$ agreement measures whether that action is ranked
among the top one or two candidates. As shown in
Table~\ref{tab:prediction_only_action_scoring}, PatientWM achieves the highest
covered top-1 and top-2 agreement, demonstrating the strongest action-planning
performance under this bounded protocol. Candidate construction, scoring, and
interpretation details are reported in Appendix~\ref{app:scorer_details}.

\begin{figure*}[!t]
  \centering
  \includegraphics[width=0.96\textwidth]{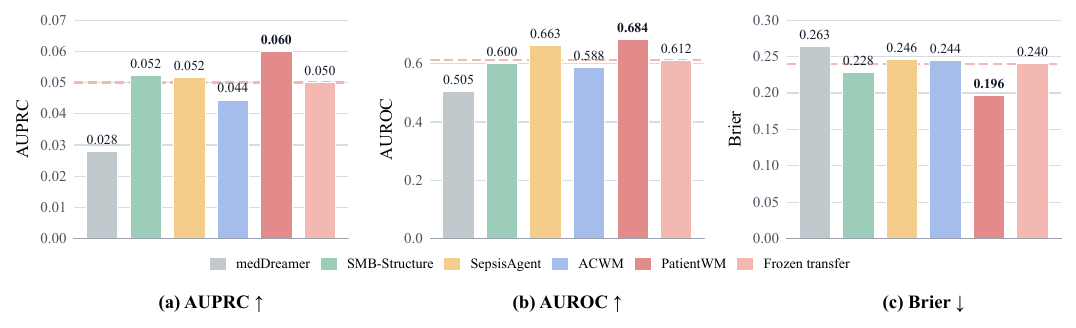}
  \caption{Frozen-state transfer on MIMIC-III Common Event-7. Panels (a)--(c)
report AUPRC, AUROC, and Brier score, respectively, with full PatientWM and
the four Table~\ref{tab:targetbenchmark} baselines shown on the same test
support. Higher is better for AUPRC and AUROC; lower is better for Brier.}
  \label{fig:frozen_state_transfer}
\end{figure*}

\subsection{Frozen-State Reuse}
We next evaluate the generalizability of the learned PatientWM state to
downstream clinical targets that are unavailable during world-model training.
To simulate this setting, we train the MIMIC-III world model without Common
Event-7 supervision, freeze all world-model parameters, and subsequently train
only a new typed decoder using the Common Event-7 labels, the learned ordered
16-state history, and decision-boundary treatment context. This protocol tests
whether a patient state learned independently of the downstream target can be
reused when a new clinical prediction task is introduced.
Figure~\ref{fig:frozen_state_transfer} compares the resulting frozen-state
transfer model with full PatientWM and the four MIMIC-III Common Event-7
baselines from Table~\ref{tab:targetbenchmark}, which are trained with the
downstream supervision, under identical target definitions, per-target
support, and evaluation. Without using Common Event-7 supervision during
world-model training, the frozen state ranks third in AUPRC and second in both
AUROC and Brier relative to the four baselines. These results indicate that the
unified patient state retains clinically relevant information beyond its
original training objectives and can serve as a reusable basis for new typed
prediction interfaces. This target-level reuse supports the central goal of
PatientWM: moving toward a clinical world model whose state is not tied to one
predefined output task. Detailed results are reported in
Appendix~\ref{app:frozen_state_reuse}.

\subsection{Label-Efficient Cross-Dataset Adaptation}

\begin{wraptable}{r}{0.52\columnwidth}
  \vspace{-1.8\baselineskip}
  \centering
  \scriptsize
  \setlength{\tabcolsep}{2.2pt}
  \renewcommand{\arraystretch}{0.96}
  \caption{MIMIC-III to MIMIC-IV adaptation with 1\% target training data.}
  \label{tab:low_data_generalization}
  \vspace{.5em}
  \resizebox{\linewidth}{!}{\begin{tabular}{lccc}
\toprule
Method & AUPRC$\uparrow$ & AUROC$\uparrow$ & Brier$\downarrow$ \\
\midrule
medDreamer       & 0.0323          & 0.4985          & 0.2204 \\
SMB-Structure    & 0.0450          & 0.5944          & 0.2061 \\
SepsisAgent      & 0.0515          & 0.5904          & 0.2035 \\
ACWM             & 0.0386          & 0.5532          & 0.2163 \\
PatientWM (Transfer) & \textbf{0.0517} & \textbf{0.6418} & \textbf{0.2023} \\
\bottomrule
\end{tabular}
}
  \vspace{-0.8\baselineskip}
\end{wraptable}

We next evaluate whether the patient state learned on MIMIC-III supports
adaptation to the MIMIC-IV Common Event-7 targets when target supervision is
limited. For each method, we freeze the source predictor and train the same
lightweight residual adapter using only 1\% of the MIMIC-IV training data. All
methods use the same target examples, outcome definitions, and evaluator. As
shown in Table~\ref{tab:low_data_generalization}, PatientWM achieves the
strongest aggregate performance across AUPRC, AUROC, and Brier score. These
results show that the frozen PatientWM state provides a transferable basis for
label-efficient adaptation to a new clinical cohort. Protocol details are
provided in Appendix~\ref{app:low_data_adaptation}.

\subsection{External Framework Evaluation}
\begin{figure}[!t]
  \centering
  \includegraphics[width=0.96\textwidth]{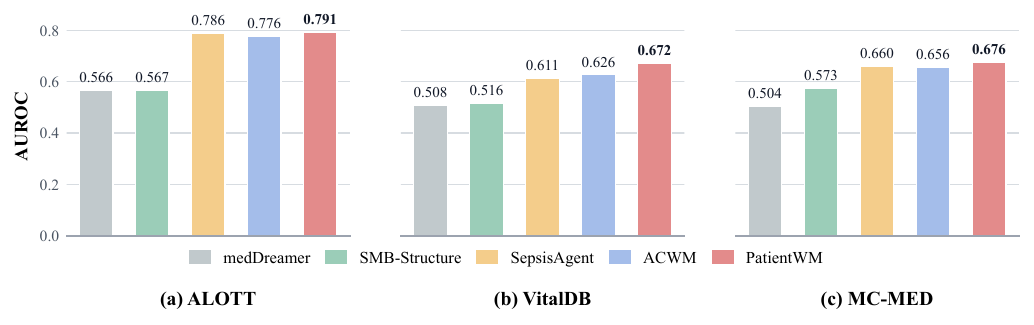}
  \caption{External AUROC comparison. Panels (a), (b), and (c) report results
  on ALOTT, VitalDB, and MC-MED, respectively. Higher is better in all panels.}
  \label{fig:external_transfer_main}
\end{figure}

To test whether PatientWM remains effective beyond the MIMIC cohorts, we
evaluate it and the baselines on three external datasets with different
clinical data objects: ALOTT~\citep{PhysioNet-alott-1.0.0},
VitalDB~\citep{lee2022vitaldb}, and
MC-MED~\citep{PhysioNet-mc-med-1.0.1}, whose
provenance and interfaces are detailed in Appendix~\ref{app:datasets}. ALOTT
and VitalDB evaluate sequential next-state prediction, whereas MC-MED provides
a pointwise prediction setting. Within each dataset, all methods use the same
visible inputs, targets, data splits, training budget, and evaluator. As shown
in Figure~\ref{fig:external_transfer_main}, PatientWM achieves the highest
AUROC on all three datasets. These results support the applicability of the
PatientWM framework across distinct clinical data interfaces; detailed AUPRC,
Brier, and AUROC results are reported in
Appendix~\ref{app:outcome_configurations}.

\section{Conclusion}

To address the limited state scope and insufficient trajectory context of
existing clinical world models, we proposed \method{}, a framework built around
a unified latent patient state and ordered state--action history. It learns the
patient state from available clinical observations, models its evolution under
clinical actions, and exposes typed clinical readouts. Experiments on MIMIC-III,
MIMIC-IV, and external datasets show strong factual prediction and recursive
rollout performance, while the action-planning evaluation shows higher
recorded-action agreement within fixed candidate sets. Together, these results
show that PatientWM learns a reusable patient state for clinical prediction,
maintains more accurate forecasts across transitions, and provides stronger
predictive support for bounded action planning.

\clearpage

\subsection*{AI use statement}
In this work, we used generative AI tools to aid and polish writing; retrieve,
discover, and synthesize related literature; support research ideation and
execution, including experiment planning, code development, result
organization and interpretation, and figure and table preparation; and draft
sections of the manuscript. The authors reviewed AI-assisted outputs against
cited sources, executable configurations, experiment logs, and aggregate
results and take responsibility for the final text, claims, code, and
artifacts.

\subsection*{Ethics statement}
This retrospective study uses deidentified, access-controlled
clinical datasets under their respective data-use terms. It does not provide
treatment recommendations or estimate causal treatment effects. Row-level
records remain within approved environments; released artifacts are limited to
code, schemas, hashes, and aggregate statistics. Appendix~\ref{app:datasets}
details data governance and intended use.

\subsection*{Reproducibility statement}
The main text reports the evaluation roles, comparison protocols,
and primary results. The appendix documents dataset construction and
information boundaries, model implementation, comparison controls,
supplementary evaluations, model parameters, and data splits.

\bibliographystyle{arxiv_conference}
\bibliography{medicalwm_references}

\clearpage
\appendix

\section[Datasets And Information Boundary]{Datasets And Information Boundary}
\label{app:datasets}

\subsection{MIMIC-III State, Outcomes, And Splits}

We align MIMIC-III clinical and waveform resources~\citep{johnson2016mimic,moody2020mimicwaveform} to contiguous six-hour ICU windows; empty windows remain in the timeline. The observed record is
\begin{equation}
  O_t=(\mathcal L_t,\mathbf m^L_t,\mathbf y_t,
  \mathbf m^y_t,\mathbf w_t,m^w_t),
\end{equation}
where $\mathcal L_t$ is a sparse set of normalized laboratory tokens, $\mathbf y_t$ contains current-window clinical events, and $\mathbf w_t\in\mathbb R^{512}$ is a frozen waveform embedding. Every component has an availability mask. A transition-derived label for $t\rightarrow t+1$ is stored in window $t+1$, never in the source window.

The resulting cohort contains 53,329 ICU episodes, 911,164 windows, and
857,835 adjacent transitions. Subject-level splitting produces
\TrainEpisodes/\ValEpisodes/\TestEpisodes train/validation/test episodes from
\TrainSubjects/\ValSubjects/\TestSubjects subjects; no subject crosses splits.
We report seven core binary outcomes with non-treatment-derived target
definitions: hypercapnic acidemia, severe acidemia, creatinine- and
urine-defined KDIGO stage 2/3 AKI, GCS drop, incident CAM positivity, and
mortality within 24 hours.

\subsection{Structured Action Ledger}

Action truth comes from \texttt{INPUTEVENTS\_CV}, \texttt{INPUTEVENTS\_MV}, and \texttt{PROCEDUREEVENTS\_MV}. Each event retains treatment identity, execution time, chart-availability time, amount, rate, unit, route when available, source table, and execution evidence. Notes may later supply decision-time context or candidate mentions, but do not override structured execution records.

We distinguish the active treatment state known at the decision boundary from
treatment records that become available later. Only boundary-known,
chart-available treatment information enters the current action port, and it
remains separate from the patient observation state. Later treatment records
are excluded from the current input and may enter only as externally supplied
logged actions at later factual-replay steps; they are not interpreted as
causal intervention assignments. Table~\ref{tab:eventsupport} reports the
dataset-wide coverage of the structured event ledger. A transition is counted
as covered when its corresponding event window contains at least one executed
event; this descriptive statistic is distinct from the availability of the
boundary-known action input.

\begin{table}[H]
  \centering
  \footnotesize
  \caption{Structured event-ledger coverage. Coverage is the fraction of transitions whose corresponding event window contains at least one executed event of any supported type. This dataset statistic does not measure boundary-known action availability or prediction performance.}
  \label{tab:eventsupport}
  \vspace{.5em}
\begin{tabular}{lrrrrr}
\toprule
Split & Transitions & Event windows & Coverage & Events & Episodes \\
\midrule
Train & 685,561 & 620,559 & 90.52\% & 11,870,214 & 41,172 \\
Validation & 85,295 & 77,490 & 90.85\% & 1,460,107 & 5,141 \\
Test & 86,979 & 78,673 & 90.45\% & 1,531,140 & 5,146 \\
\bottomrule
\end{tabular}

\end{table}

\subsection{MIMIC-IV Port}

We use MIMIC-IV for two principal native-cohort evaluation roles under the same
six-hour transition unit and decision-time information boundary. The first is
the cross-dataset Common Event-7 clinical readout reported in
Table~\ref{tab:targetbenchmark},
which shares its target definitions with the corresponding MIMIC-III panel.
The second trains a MIMIC-IV transition model used for the multi-step rollout
in Figure~\ref{fig:mimiciv_strict_rollout} and the action-planning evaluation
in Table~\ref{tab:prediction_only_action_scoring}. The frozen transition also
supports the MIMIC-IV typed laboratory decoder reported in
Table~\ref{tab:mimiciv_matched_test}. These roles share the dataset
construction, patient-disjoint split, availability-mask and leakage contracts.
Separately, the label-efficient adaptation experiment uses MIMIC-IV as the
target cohort for evaluating transfer from a frozen MIMIC-III source predictor.

\subsection{Common Event-7 Panel}

To support a direct cross-dataset event comparison, we derive a separate
Common Event-7 outcome sidecar from the native query matrices in both datasets.
The shared heads are any acute deterioration, new invasive ventilation, new
pressor, new circulation pressor, respiratory IMV escalation, Sepsis-3 onset,
and dynamic ICU mortality. The first six heads are next-window events; dynamic
ICU mortality retains its six-hour gap and 24-hour target horizon. Each row
keeps the observed mask and native target timing. The sidecar is used for
task-level readout evaluation and is not inserted into the six-dimensional
recursive state label matrix. The MIMIC-III Core-7 and MIMIC-IV cardiac-arrest
extensions remain dataset-specific transition diagnostics.

\subsection{External Cohorts}

The supplementary question is whether the same unified-patient-state framework
applies across different clinical data objects.
ALOTT~\citep{PhysioNet-alott-1.0.0} and
VitalDB~\citep{lee2022vitaldb} have verified adjacent states;
MC-MED~\citep{PhysioNet-mc-med-1.0.1} has one prediction anchor and serves as a
pointwise control. Before
task eligibility filtering, their train/test counts are
974,025/215,239, 881,421/111,162, and 227,886/33,119, respectively. VitalDB
aggregates 22 forecasting tasks and excludes three concurrent-state controls.
One MC-MED task with only three positive test cases after waveform QC is
excluded from the aggregate. Every method receives pre-anchor clinical
context and a known-empty executed-action port.

\subsection{Data Governance And Intended Use}

MIMIC-III is a deidentified, credentialed PhysioNet resource governed by its data-use agreement~\citep{johnson2016mimic}; the waveform resource is separately released through PhysioNet~\citep{moody2020mimicwaveform}. MIMIC-IV is separately governed and deidentified~\citep{johnson2023mimic}. VitalDB is used under the terms of the versioned release~\citep{lee2022vitaldb}. ALOTT and MC-MED are used from their versioned PhysioNet releases~\citep{PhysioNet-alott-1.0.0,PhysioNet-mc-med-1.0.1}. Public artifacts contain code, schemas, hashes, and aggregate statistics only; governed row-level trajectories remain inside their approved environments.

\method is intended for retrospective representation and factual-forecasting research, not bedside monitoring, treatment selection, or autonomous simulation. Clinical use would require external and prospective validation, subgroup and calibration audits, and a causal design appropriate to the intended decision.

\section[PatientWM Method And Implementation]{PatientWM Method And Implementation}

\subsection{Architecture}

\paragraph{Observation Encoder.}
The main MIMIC-III checkpoint instantiates $E_O$ with three active observation
branches: sparse laboratory tokens, masked clinical facts, and frozen
waveform-derived embeddings. Each branch emits a 256-dimensional modality
representation, and availability masks determine which branch outputs are
eligible for fusion.

Sparse laboratory observations are represented as at most 128 valid item
tokens per window from a 693-item lab vocabulary. For item $j$, the encoder
combines a learned identity embedding $e_j^L\in\mathbb R^{256}$ with an
11-dimensional numeric feature vector $x_{t,j}^L$ containing normalized value
summaries, observation counts, abnormality summaries, timing or recency, and
missingness-related indicators. The token representation is
\begin{equation}
  h_{t,j}^L
  =
  \operatorname{LayerNorm}\!\left(e_j^L+W_Lx_{t,j}^L+b_L\right).
  \label{eq:appendix_lab_token_encoder}
\end{equation}
A learned lab query attends over valid lab tokens using the lab key-padding
mask. If no valid lab token exists in the window, the branch returns a learned
empty-lab embedding and marks the lab modality absent for fusion.

The clinical-fact branch consumes six binary current-window facts with their
availability masks. Unavailable fact values are masked before projection, and
the concatenated values and masks are mapped to a 256-dimensional fact
representation:
\begin{equation}
  h_t^y
  =
  \operatorname{LayerNorm}
  \left(
    W_y[\mathbf y_t\odot\mathbf m_t^y;\mathbf m_t^y]+b_y
  \right).
  \label{eq:appendix_fact_encoder}
\end{equation}
The fact modality is marked present when at least one fact is observed.
The world model does not ingest raw waveform samples. A frozen pretrained
waveform encoder, without downstream fine-tuning, converts valid waveform
frames into embeddings, which are mean-pooled into one 512-dimensional
representation per window. When this representation is present, a linear
projection maps it to the common 256-dimensional branch width; when absent, a
learned empty-waveform embedding is used and the waveform modality is marked
absent.

For modality fusion, each branch representation receives a learned modality
type embedding. A learned observation query attends over the stack of available
branch states:
\begin{equation}
  \latent_t
  =
  E_O(O_t)
  =
  \operatorname{LayerNorm}
  \left(
    \operatorname{MHA}
    (q_O,G_t^{\mathrm{mod}},G_t^{\mathrm{mod}};
     \mathbf m_t^{\mathrm{mod}})
  \right),
  \label{eq:appendix_modality_fusion}
\end{equation}
where $G_t^{\mathrm{mod}}\in\mathbb R^{3\times256}$ stacks the lab, fact, and
waveform branch states, and $\mathbf m_t^{\mathrm{mod}}$ records branch
availability. If all branches are absent, the encoder returns a learned
empty-state embedding.

\paragraph{Online And Target Encoders.}
Training uses an online observation encoder $E_O$ and a momentum target encoder
$\bar E_O$. The target encoder supplies the stopped-gradient next-window
latent target,
\begin{equation}
  \latent_{t+1}^{\mathrm{tar}}
  =
  \operatorname{stopgrad}\!\left(\bar E_O(O_{t+1})\right),
  \label{eq:appendix_target_encoder}
\end{equation}
and its parameters follow the online encoder after each optimizer update:
\begin{equation}
  \bar\phi\leftarrow0.99\,\bar\phi+0.01\,\phi.
  \label{eq:appendix_target_encoder_ema}
\end{equation}
This target is a joint-embedding predictive supervision signal. It is not
interpreted as recovered true latent physiology.

\paragraph{Action Encoder.}
The main checkpoint encodes the source-boundary active-treatment record from the
structured action ledger. A treatment item is included in $\action_t$ only when
it started before the end of window $t$, remains active at that
boundary, and is already chart-available. Target-window duration, future
delivered amount, future start or stop, end-of-window activity, and future
status changes are zeroed or excluded.

The active-treatment set contains at most 64 tokens from a 308-entry treatment
vocabulary. Each token has a 25-dimensional numeric feature vector covering
available lifecycle, exposure, amount/rate/unit/source, execution-evidence, and
timing features. Let $e_j^A\in\mathbb R^{256}$ be the treatment identity
embedding and $x_{t,j}^A\in\mathbb R^{25}$ its numeric feature vector:
\begin{equation}
  h_{t,j}^A
  =
  \operatorname{LayerNorm}
  \left(e_j^A+W_Ax_{t,j}^A+b_A\right).
  \label{eq:appendix_action_token_encoder}
\end{equation}
A learned action query attends over the valid action tokens:
\begin{equation}
  \boldsymbol{\alpha}_t
  =
  E_A(\action_t)
  =
  \operatorname{LayerNorm}
  \left(
    \operatorname{MHA}
    (q_A,G_t^A,G_t^A;\mathbf m_t^A)
  \right)
  \in\mathbb R^{256},
  \label{eq:appendix_action_set_encoder}
\end{equation}
where $G_t^A$ stacks token states and $\mathbf m_t^A$ masks invalid token
slots. A known-empty action ledger returns a learned empty-action embedding and
an explicit presence indicator; it is not treated as missing data.

\paragraph{Transition Model.}
The main transition first projects the encoded patient state and encoded action
state into a transition token:
\begin{equation}
  x_t^{\mathrm{tr}}
  =
  \operatorname{LayerNorm}
  \left(
    \operatorname{GELU}
    \left(
      W_{\mathrm{tr}}[\latent_t;\boldsymbol{\alpha}_t]
      +b_{\mathrm{tr}}
    \right)
  \right)
  \in\mathbb R^{256}.
  \label{eq:appendix_transition_token}
\end{equation}

The primary dynamics operator is a pre-norm causal Transformer~\citep{vaswani2017attention}. It uses strict
causal masking and a transition-valid padding mask so that position $t$ attends
only to legal positions through $t$.
The architecture and training settings are collected in Table~\ref{tab:implementation_settings}.

For final hidden state $h_t\in\mathbb R^{256}$,
\begin{equation}
  h_t
  =
  \operatorname{CausalTransformer}
  \left(x_{t-C+1:t}^{\mathrm{tr}}\right).
  \label{eq:appendix_transformer_hidden}
\end{equation}
The latent transition is decoded as a gated residual around the current latent
state:
\begin{equation}
  \Delta_t^z=W_\Delta^zh_t+b_\Delta^z,
  \qquad
  g_t^z=\sigma(W_g^zh_t+b_g^z),
  \qquad
  \widehat\latent_{t+1}
  =\latent_t+g_t^z\odot\Delta_t^z.
  \label{eq:appendix_residual_transition}
\end{equation}
The residual projection is zero-initialized, and the gate bias initializes the
gate probability near 0.1. The initialized model is therefore close to latent
copy-forward and learns departures only when supported by training targets.

The main MIMIC-III configuration optimizes 5,685,029 parameters. The momentum
target encoder follows the online encoder by exponential moving average and is
not optimized directly.

\paragraph{Typed Readout Interface.}
Following Equation~\ref{eq:typed_observation_readout}, each type-specific
decoder maps the predicted patient state $\widehat\latent_{t+1}$ to its
corresponding clinical target. Continuous heads may additionally use the
transition representation or an output-specific causal prior. These remain
decoder-side inputs rather than additional shared patient states, and the
recursively reused patient-state slot remains $\widehat\latent_{t+1}$.

\subsection{Typed Decoders And Recursive State}

\paragraph{Persistence-Aware Continuous Readouts.}
As described in Equation~\ref{eq:persistence_aware_readout}, continuous
readouts predict a gated residual around a causal persistence prior:
\begin{equation}
  \widehat r_{t+1}
  =
  r_t^{\mathrm{prior}}
  +
  g_t^r\odot\Delta_t^r .
  \label{eq:appendix_generic_residual_head}
\end{equation}
For waveform embeddings, the prior is the most recent quality-qualified
embedding, with a learned fallback when none is available. For laboratory
values, the prior is the last observed normalized value for a previously
measured item and the training-set mean for a never-observed item. These priors
use only source-visible information and remain within their corresponding
decoders during rollout.

\paragraph{Decoder Roles.}
Each query type selects a decoder with a target-specific output space. The
transition checkpoint also includes auxiliary heads that ground the learned
state in next-window observations. Table~\ref{tab:appendix_decoder_routing}
summarizes these roles.
\begin{table}[h]
  \centering
  \footnotesize
  \caption{Typed decoder outputs and their roles in training and evaluation.}
  \label{tab:appendix_decoder_routing}
  \vspace{.5em}
  \resizebox{\columnwidth}{!}{%
  \begin{tabular}{lll}
    \toprule
    Decoder & Output & Role \\
    \midrule
    clinical event/outcome & target-specific probability & clinical readout \\
    waveform & 512-D embedding & waveform prediction and cache update \\
    clinical fact & six probabilities & transition supervision \\
    lab availability & 693 probabilities & observation-process supervision \\
    joint lab value & 693 normalized values & transition supervision and cache update \\
    queried lab value & one normalized value & observed-laboratory prediction \\
    \bottomrule
  \end{tabular}
  }
\end{table}

\paragraph{Clinical And Waveform Readouts.}
The query type selects the corresponding decoder. Binary clinical-event and
outcome decoders emit probabilities for their requested targets according to
Equation~\ref{eq:typed_observation_readout}. These task-specific readouts share
the predicted patient state but use separate output parameters and losses.

The transition checkpoint additionally predicts six next-window clinical facts
with a linear logit head,
\begin{equation}
  \widehat{\mathbf p}_{t+1}^{y}
  =
  \sigma(W_y^{\mathrm{dec}}h_t+b_y^{\mathrm{dec}}).
  \label{eq:appendix_fact_decoder}
\end{equation}
These auxiliary factual targets ground transition training and are distinct
from the Core-7 and Common Event-7 clinical readout panels used in the main
evaluation.

The waveform decoder predicts a residual in the frozen embedding space:
\begin{equation}
  \widehat{\mathbf w}_{t+1}
  =
  \mathbf w_t^{\mathrm{prior}}
  +
  g_t^w\odot\Delta_t^w,
  \qquad
  \Delta_t^w=W_\Delta^wh_t+b_\Delta^w,
  \qquad
  g_t^w=\sigma(W_g^wh_t+b_g^w).
  \label{eq:appendix_waveform_decoder}
\end{equation}
If the waveform prior is unavailable, a separate linear fallback maps $h_t$ to
the 512-dimensional embedding. The decoded waveform is a model-generated
embedding-space representation, not raw waveform generation.

\paragraph{Joint Laboratory Heads.}
The laboratory-availability decoder emits one next-window observation
probability per item in the 693-item vocabulary:
\begin{equation}
  \widehat{\mathbf p}_{t+1}^{L,\mathrm{present}}
  =
  \sigma(W_m^Lh_t+b_m^L)
  \in[0,1]^{693}.
  \label{eq:appendix_lab_presence_decoder}
\end{equation}
The corresponding dense value head predicts normalized laboratory values
around the causal prior:
\begin{equation}
  \widehat{\mathbf l}_{t+1}^{\mathrm{joint}}
  =
  \mathbf l_t^{\mathrm{prior}}
  +
  \mathbf g_t^L\odot\Delta_t^L .
  \label{eq:appendix_joint_lab_value_decoder}
\end{equation}
This head contributes loss only for numerically observed target item-value
pairs. During recursive rollout, its predictions update the cache for items
that have already been observed; the availability probabilities remain typed
outputs and do not require thresholding to create new synthetic measurements.

\paragraph{Accumulated Laboratory Decoder.}
For the MIMIC-III observed-laboratory evaluation, the trained transition is
frozen and only a query-specific laboratory decoder is optimized. The decoder
receives the rolled patient state or transition context, the queried item
identity, its last source-visible normalized value and age, whether it has been
observed previously, and causal summary features computed from the available
laboratory history.

Its GapShrink prior is
\begin{equation}
  \widehat l_{j,t+h}^{\mathrm{gap}}
  =
  \beta_{j,a_j+h}l_{j,t}^{\mathrm{last}},
  \qquad
  \beta_{j,g}\in[0,1],
  \label{eq:appendix_gapshrink_prior}
\end{equation}
where $a_j$ is the age of the last observed value at rollout start. The
coefficients are estimated from training data only, and gaps beyond the modeled
range fall back to persistence.

The learned query branch combines the 256-dimensional rolled state, a
128-dimensional item embedding, and the source-visible laboratory ledger into
a 128-dimensional query representation $q_{j,h}$. Cross-laboratory attention
allows queried item $j$ to use measurements of other previously observed
items:
\begin{equation}
  A_{jk}
  =
  \operatorname{softmax}_k
  \left(
    \frac{(W_Qq_{j,h})^\top(W_Ks_k)}{\sqrt{128}}
  \right),
  \qquad
  c_j^{\mathrm{cross}}
  =
  \sum_{k:\mathrm{seen}} A_{jk}W_Vs_k .
  \label{eq:appendix_cross_lab_attention}
\end{equation}
The attended context is projected back into the query representation,
$\widetilde q_{j,h}=q_{j,h}+W_Oc_j^{\mathrm{cross}}$. The output projection
$W_O$ is zero-initialized so that training begins from the item-specific query
model and adds cross-item corrections only when supported by the data. The
final prediction is a gated residual around GapShrink:
\begin{equation}
  \delta_{j,h}=W_\delta \widetilde q_{j,h}+b_\delta,
  \qquad
  g_{j,h}=\sigma(W_g\widetilde q_{j,h}+b_g),
  \qquad
  \widehat l_{j,t+h}
  =
  \widehat l_{j,t+h}^{\mathrm{gap}}
  +
  g_{j,h}\delta_{j,h}.
  \label{eq:appendix_accumulated_lab_decoder}
\end{equation}
The gate bias initializes the update probability near 0.1, and the decoder is
optimized with pointwise $L_1$ loss on observed item--target pairs. For each
item, the accumulated source-only features comprise the running mean, running
standard deviation, deviation of the latest value from the running mean, fast
and slow exponentially weighted moving averages, their difference, and
positive and negative CUSUM statistics. The fast and slow coefficients are 0.5
and 0.1, respectively; CUSUM uses retention 0.85 and drift 0.05 with gap-aware
decay on the six-hour grid. These statistics are computed only from
source-visible observations and are never refreshed from the future target
ledger.

\paragraph{Recursive Implementation State.}
Recursive evaluation reuses the predicted patient state together with the
causal continuous-readout caches:
\begin{equation}
  R_t=
  \left(
    \widehat\latent_t,
    \widehat{\mathbf w}_t,
    \mathbf c_t^L,
    \mathbf s_t^L,
    \mathbf a_t^L
  \right),
  \label{eq:appendix_recursive_state}
\end{equation}
where $\widehat\latent_t$ is the predicted latent patient state,
$\widehat{\mathbf w}_t$ is the generated waveform-embedding cache,
$\mathbf c_t^L$ is the dense causal last-value lab cache, $\mathbf s_t^L$
records whether each lab item has ever been seen, and $\mathbf a_t^L$ records
lab-value age in windows. The additional fields support typed persistence and
query decoding; $\latent_t$ remains the unified patient state used by the
transition model. Clinical-event, outcome, and laboratory-availability
probabilities are decoded from the evolving state when requested, whereas the
continuous caches above are carried to provide causal priors for subsequent
predictions.

\subsection{Training Details}

\paragraph{Direct Transition Objective.}
For a valid transition $t\rightarrow t+1$, PatientWM jointly optimizes the
latent transition and the available typed targets:
\begin{equation}
  \mathcal L_{\mathrm{direct}}
  =
  \lambda_z\mathcal L_z
  +\lambda_w\mathcal L_w
  +\lambda_y\mathcal L_y
  +\lambda_{m^L}\mathcal L_{m^L}
  +\lambda_L\mathcal L_L .
  \label{eq:appendix_direct_loss_decomposition}
\end{equation}
The five terms correspond to the next patient state, waveform embedding,
clinical facts, laboratory availability, and observed laboratory values. In
the reported transition configurations, every enabled top-level term has
weight 1.0; unavailable targets are removed by their corresponding masks.

The latent objective aligns the predicted patient state with the
stopped-gradient target from the momentum encoder:
\begin{equation}
  \mathcal L_z
  =
  1-\cos(\widehat\latent_{t+1},\latent_{t+1}^{\mathrm{tar}}).
  \label{eq:appendix_latent_loss}
\end{equation}
The waveform objective is evaluated only when a quality-qualified next-window
embedding is available:
\begin{equation}
  \mathcal L_w
  =
  \mathbf 1[m_{t+1}^{w}=1]
  \left(
    1-\cos(\widehat{\mathbf w}_{t+1},\mathbf w_{t+1})
  \right).
  \label{eq:appendix_waveform_loss}
\end{equation}
Clinical-fact and laboratory-availability heads use masked binary
cross-entropy. Their positive weights are estimated from the training split as
negative-to-positive ratios and clipped to $[1,100]$. The joint laboratory
value head uses masked Smooth-$L_1$ loss only for item--target pairs with an
observed numeric value.

\paragraph{Residual-Update Regularization.}
The magnitudes of gated latent, waveform, and laboratory updates are
regularized by
\begin{equation}
  \mathcal L_{\mathrm{resid}}
  =
  \frac{1}{3}\left(
    \|g^z\odot\Delta^z\|_2^2
    +\|g^w\odot\Delta^w\|_2^2
    +\|g^L\odot\Delta^L\|_2^2
  \right).
  \label{eq:appendix_residual_update_regularizer}
\end{equation}
This term discourages unsupported departures from the causal persistence
priors while allowing the learned gates to apply larger updates when supported
by the training targets.

\paragraph{Recursive Training Objective.}
Transition models trained for multi-step use combine the teacher-forced direct
objective with a generated-history branch:
\begin{equation}
  \mathcal L
  =
  \mathcal L_{\mathrm{direct}}
  +\lambda_r(t)
  \sum_{k=1}^{H}
  \mathcal L_{\mathrm{direct}}^{(t+k\mid t)}
  +\lambda_\Delta\mathcal L_{\mathrm{resid}}
  +\mathcal L_{\mathrm{aux}}.
  \label{eq:appendix_closed_loop_training_loss}
\end{equation}
The generated-history branch receives the observed initial state once and then
advances the predicted latent state, waveform embedding, and existing-item
laboratory cache. Logged source-boundary actions remain externally supplied at
each transition. Future observations, values, and availability masks enter
only as supervised targets. The rollout coefficient $\lambda_r(t)$ is warmed
up from zero to its final value in the MIMIC-III transition training protocol,
avoiding long generated prefixes at the beginning of optimization.

\begin{table}[h]
  \centering
  \footnotesize
  \caption{Recursive training settings used by the reported transition experiments.}
  \label{tab:appendix_recursive_training_settings}
  \vspace{.5em}
  \begin{tabular}{lccc}
    \toprule
    Experiment & $H$ & Final $\lambda_r$ & Additional objective \\
    \midrule
    MIMIC-III transition rollout & 16 & 0.25 & residual and waveform terms \\
    MIMIC-IV primary rollout & 16 & 0.50 & residual term \\
    One-step/readout training & -- & 0 & none \\
    \bottomrule
  \end{tabular}
\end{table}
For recursive transition training, $\lambda_\Delta=0.001$.
Modality-specific auxiliary losses are included only when the corresponding
target is part of the stated experiment.

\paragraph{Teacher-Forced And Free-Running Modes.}
Teacher-forced evaluation constructs $O_t$ and $\action_t$ from the legal
observed record and uses the observed prefix to form the ordered state--action
history. The next record $O_{t+1}$ is joined only to provide targets. This mode
therefore measures factual next-window prediction without testing error
accumulation under generated history.

Logged-action free-running evaluation initializes once from observed
$O_{t_0}$. At each subsequent step, it encodes the externally supplied logged
action, applies the transition to the retained predicted state--action prefix,
updates the latent, waveform, and existing-item laboratory caches, and decodes
the requested outputs. The observed future record is joined only after the
prediction for scoring. The principal reported horizons of 1, 4, 8, and 16
transitions correspond to 6, 24, 48, and 96 hours, respectively.

\subsection{Hyperparameters}

\begin{table}[H]
  \centering
  \footnotesize
  \caption{Implementation settings for the main MIMIC-III causal-Transformer transition and laboratory readout.}
  \label{tab:implementation_settings}
  \label{tab:appendix_transition_hyperparams}
  \label{tab:appendix_main_training_settings}
  \label{tab:appendix_v95_settings}
  \vspace{.5em}
  \textbf{Transition Architecture}\par\smallskip
\resizebox{0.84\textwidth}{!}{\begin{tabular}{ll}
    \toprule
    Setting & Value \\
    \midrule
    Width & 256 \\
    Layers & 4 \\
    Attention heads & 8 \\
    Feed-forward hidden width & 1,024 \\
    Activation & GELU \\
    Dropout & 0.1 \\
    Maximum context & 16 six-hour windows \\
    Positional representation & learned absolute position embedding \\
    Causal mask & strict upper-triangular attention mask \\
    \bottomrule
  \end{tabular}}

\par\medskip
\textbf{Transition Optimization}\par\smallskip
\resizebox{0.84\textwidth}{!}{\begin{tabular}{ll}
    \toprule
    Setting & Value \\
    \midrule
    Optimizer & AdamW \\
    Learning rate & $3\times10^{-4}$ \\
    Weight decay & 0.01 \\
    Micro-batch & 8 \\
    Gradient accumulation & 4 \\
    Effective batch & 32 \\
    Optimizer steps & 1,024 \\
    Balanced shard contribution & 32 optimizer-step equivalents per shard \\
    Sampled trajectories & 32,768 \\
    Checkpoint criterion & direct validation loss plus 0.25 closed-loop loss \\
    \bottomrule
  \end{tabular}}

\par\medskip
\textbf{Query-Only Laboratory Readout}\par\smallskip
\resizebox{0.84\textwidth}{!}{\begin{tabular}{ll}
    \toprule
    Setting & Value \\
    \midrule
    Complete frozen-core-plus-decoder parameters & 9,132,456 \\
    Trainable decoder parameters & 191,107 \\
    Item-query dimension & 128 \\
    Training mode & recursive rollout \\
    Point loss & micro-target masked $L_1$ \\
    NLL weight & 0 \\
    Batch size & 8 \\
    Gradient accumulation & 4 \\
    Learning rate & $3\times10^{-4}$ \\
    Weight decay & 0.01 \\
    Optimizer steps & 2,048 \\
    Examples seen & 65,536 across all 32 shards \\
    \bottomrule
  \end{tabular}}

\end{table}

\subsection{Tensor Contract}

The following contract describes the main MIMIC-III transition implementation.
The MIMIC-IV implementations preserve the same temporal alignment, masking, and
source--target separation while using their dataset-specific observation and
target schemas. Let $B$ be the batch size, $C\le16$ the number of transitions,
$N_L=128$ the maximum number of laboratory tokens per window, $N_A=64$ the
maximum number of active-treatment tokens, $V_L=693$ the laboratory
vocabulary size, $K_y=6$ the auxiliary clinical-fact dimension, $D_W=512$ the
waveform-embedding dimension, and $D=256$ the patient-state dimension.

\begin{table*}[t]
  \centering
  \footnotesize
  \caption{Tensor contract for the main MIMIC-III transition model. The state
  axis contains $C+1$ windows because $C$ transitions connect consecutive
  source and target states.}
  \label{tab:appendix_tensor_contract}
  \vspace{.5em}
  \resizebox{0.96\textwidth}{!}{%
  \begin{tabular}{lll}
    \toprule
    Tensor role & Shape & Meaning \\
    \midrule
    laboratory item identities & $[B,C+1,N_L]$ & sparse item indices for each state window \\
    laboratory numeric features & $[B,C+1,N_L,11]$ & value summaries, counts, timing, and availability features \\
    laboratory token mask & $[B,C+1,N_L]$ & valid, non-padding laboratory positions \\
    clinical-fact values & $[B,C+1,K_y]$ & current-window binary clinical facts \\
    clinical-fact mask & $[B,C+1,K_y]$ & availability of each clinical fact \\
    waveform embedding & $[B,C+1,D_W]$ & frozen waveform-derived representation \\
    waveform mask & $[B,C+1]$ & availability of the waveform representation \\
    \midrule
    treatment item identities & $[B,C,N_A]$ & source-boundary active-treatment indices \\
    treatment numeric features & $[B,C,N_A,25]$ & lifecycle, exposure, amount, evidence, and timing features \\
    treatment token mask & $[B,C,N_A]$ & valid, non-padding treatment positions \\
    transition-valid mask & $[B,C]$ & legal, non-padding source--target transitions \\
    \midrule
    encoded patient states $\latent$ & $[B,C+1,D]$ & observation-derived state sequence \\
    encoded actions $\boldsymbol{\alpha}$ & $[B,C,D]$ & treatment-set representations \\
    transition hidden states $h$ & $[B,C,D]$ & causal history-conditioned representations \\
    predicted patient states $\widehat\latent$ & $[B,C,D]$ & predictions aligned with state positions $2{:}C+1$ \\
    predicted waveform embeddings & $[B,C,D_W]$ & next-window waveform representations \\
    clinical-fact logits & $[B,C,K_y]$ & auxiliary next-window fact predictions \\
    laboratory-availability logits & $[B,C,V_L]$ & next-window item-observation predictions \\
    laboratory-value predictions & $[B,C,V_L]$ & dense normalized next-window values \\
    \bottomrule
  \end{tabular}
  }
\end{table*}

For transition index $c$, the inputs at state position $c$ and action position
$c$ predict targets at state position $c+1$. Padding masks are applied both
within each sparse set and across the transition sequence; unavailable values
are masked before encoding and do not contribute to their typed losses. Longer
histories retain the most recent $C$ valid transitions. During recursive
evaluation, the observed tensors at future state positions remain scoring
targets: after the initial window, the transition consumes the predicted
patient state and causal continuous caches together with the externally
supplied action sequence.

\section{Comparison Methods And Controls}
\label{app:comparison_details}

\paragraph{Why Adaptation Is Necessary.}
The source methods were developed for different clinical data interfaces and
prediction objectives. We adapt them to a common clinical prediction interface
to compare how their retained dynamics or representation structures model
patient evolution from matched information and supervision.

\paragraph{Adaptation Principle And Fairness.}
For each comparison, we retain the defining computational core of the source
method and replace the dataset-facing input and output interfaces. Methods are
matched on patient splits, temporal windows, source-visible inputs, action
timing, targets, missingness masks, eligible support, typed readouts, losses,
optimization budget, validation-based selection, and test evaluation.
Parameter counts and internal recurrence remain properties of the retained
method structures.

\paragraph{MIMIC-IV Comparisons.}
Table~\ref{tab:mimiciv_adapter_contract} summarizes the MIMIC-IV comparison
contract. SepsisAgent, ACWM, and medDreamer are retrained on the common
MIMIC-IV tensors after their native input and output layers are replaced by
shared laboratory and clinical-fact encoders, the same frozen waveform
encoder, the common action port, and typed readouts. Their recurrent or latent dynamics
blocks retain the source-inspired update structure. SMB-Structure receives
the same source-visible modality representations; its frozen predictive core
is followed by an action-aware transition and the shared typed readouts for
the rollout and action-planning experiments.

\begin{table*}[htbp]
  \centering
  \footnotesize
  \caption{Adaptation and fairness contract for the MIMIC-IV comparisons.
  Each method retains its defining source-inspired core while using the stated
  common benchmark interface.}
  \label{tab:mimiciv_adapter_contract}
  \vspace{.5em}
  \begin{tabular}{@{}p{0.11\textwidth}p{0.21\textwidth}p{0.28\textwidth}p{0.32\textwidth}@{}}
\toprule
Method & Retained core & Interface adaptation & Fairness controls \\
\midrule
medDreamer & Dreamer-style recurrent state-space dynamics & Replace task-specific interfaces with shared laboratory and clinical-fact encoders, frozen waveform embeddings, action inputs, and typed readouts & Same split, windows, visible inputs, action timing, targets, masks, training budget, selection rule, support, and evaluator \\
SMB-Structure & Frozen predictive representation & Map the same laboratory, clinical-fact, and frozen-waveform representations into the frozen core, then fit an action-aware transition and shared typed readouts & Same split, visible inputs, action timing, clinical targets, masks, support, and evaluator \\
SepsisAgent & Recurrent history-to-next-state update & Replace native feature and output layers with shared laboratory and clinical-fact encoders, frozen waveform embeddings, action inputs, and typed readouts & Same split, windows, visible inputs, action timing, targets, masks, training budget, selection rule, support, and evaluator \\
ACWM & Latent state-transition update & Replace native signal/label interface with shared laboratory and clinical-fact encoders, frozen waveform embeddings, action inputs, and typed readouts & Same split, windows, visible inputs, action timing, targets, masks, training budget, selection rule, support, and evaluator \\
\bottomrule
\end{tabular}

\end{table*}

\paragraph{MIMIC-III Target-Specific Controls.}
MIMIC-III uses references appropriate to each predicted object rather than one
pooled baseline. Waveform-embedding prediction is compared with copy-forward
and state-only, aligned-action, and shuffled-action controls. Observed
laboratory values are compared with last-value carry-forward and the
training-only GapShrink prior. Clinical outcomes use train prevalence and
recurrent clinical-prediction baselines evaluated with identical target masks.
Laboratory availability uses mask-copy or training prevalence where
applicable. Each comparison therefore preserves the target definition,
eligible support, and metric used by PatientWM.

\paragraph{Component And Modality Ablations.}
The five Core-7 ablations in Table~\ref{tab:t004_module_ablation} use the same
MIMIC-III split, outcome definitions, readout architecture, optimization
budget, checkpoint selection rule, and evaluator as the full configuration.
The first group removes action conditioning or ordered state--action history. The
second group restricts the observation encoder to laboratory values, clinical
facts, or waveform embeddings. These controls isolate transition mechanisms
and modality composition from changes in the downstream prediction task.

\paragraph{External-Cohort Adaptations.}
The same adaptation rule is applied independently within ALOTT, VitalDB, and
MC-MED. Dataset-specific encoders map the available variables into a common
state width, after which each method applies its retained dynamics core and the
same target-specific output heads. Within a dataset, methods share training and
validation splits, visible inputs, target definitions, losses, optimization
budget, checkpoint rule, and evaluator. ALOTT and VitalDB provide sequential
examples. MC-MED has no verified adjacent-state sequence, so all temporal
methods operate with context length one. None of these external interfaces
provides a verified executed-action update; the action port is therefore
explicitly known-empty for every method.

\section[Experimental Protocol And Supplementary Results]{Experimental Protocol And Supplementary Results}
\subsection{Evaluation Protocol}
\label{app:evaluation_protocol}

\paragraph{Evaluation Roles.}
The experiments separate four roles that answer different questions. Factual
transition evaluation measures next-state prediction along recorded
trajectories. Typed readout evaluation measures requested clinical events,
outcomes, and continuous values from the learned state. Recursive rollout
evaluation repeatedly advances generated states without consuming future
observations. The action-planning evaluation passes a bounded set of externally
supplied candidates through the same transition model and measures agreement
with the recorded clinician action.

\paragraph{Temporal Evaluation Modes.}
Teacher-forced evaluation encodes each legal observed source state and scores
the immediately following target. Logged-action free-running evaluation
observes the initial prefix once, then reuses generated patient states and
causal caches while supplying the recorded source-boundary action externally
at every transition. The principal rollout tables report 1, 4, 8, and 16
six-hour transitions, corresponding to 6, 24, 48, and 96 hours. Future
observations, values, and availability masks are joined after prediction for
scoring. Generated states and causal-cache entries remain model outputs and are
not relabeled as observed clinical records. Aligned, removed, and
patient-shuffled action controls change only the action input and probe factual
action dependence.

\paragraph{Target-Specific Metrics.}
Average precision (AP) is the empirical precision--recall area used here as
AUPRC. For target $q$, after sorting its $N_q$ predictions by decreasing
score, we compute
\begin{equation}
  \operatorname{AP}_q
  =
  \sum_{r=1}^{N_q}
  \left(R_{q,r}-R_{q,r-1}\right)P_{q,r},
  \qquad
  \operatorname{MacroAUPRC}
  =
  \frac{1}{|\mathcal Q_{\mathrm{eval}}|}
  \sum_{q\in\mathcal Q_{\mathrm{eval}}}\operatorname{AP}_q,
  \label{eq:appendix_macro_auprc}
\end{equation}
where $P_{q,r}$ and $R_{q,r}$ are precision and recall at rank $r$, and
$\mathcal Q_{\mathrm{eval}}$ contains heads with valid evaluation support.
Macro-AUROC uses the same unweighted aggregation over per-target AUROC. AUPRC
is emphasized for imbalanced binary targets, while AUROC measures case--control
ranking~\citep{davis2006relationship,saito2015precision,harutyunyan2019multitask}.

For predicted probability $p_{i,q}$ and binary target $y_{i,q}$, probability
accuracy is measured by
\begin{equation}
  \operatorname{Brier}_q
  =
  \frac{1}{N_q}
  \sum_{i=1}^{N_q}
  \left(p_{i,q}-y_{i,q}\right)^2,
  \label{eq:appendix_brier}
\end{equation}
with lower values indicating better probabilistic accuracy
\citep{brier1950verification}.

The Common Event-7 interface contains six shared acute next-window events and
dynamic ICU mortality: acute deterioration, new invasive ventilation, acute
pressor initiation, circulatory pressor initiation, respiratory IMV
escalation, Sepsis-3 onset, and mortality. Mortality retains its native
six-hour gap and 24-hour prediction horizon. Dataset-specific transition heads
are not included in this cross-dataset macro. The per-target metrics use each
head's observed support before the seven values are macro-averaged.

Observed-laboratory nMAE is absolute error after item-wise normalization using
training statistics and is evaluated only for item--target pairs measured in
the future. This avoids aggregating errors across incompatible physical
scales~\citep{hyndman2006another}. Let $\mu_j$ and $\sigma_j$ be the
training-set mean and scale for laboratory item $j$. Over the observed future
item--target set $\Omega_h$ at horizon $h$,
\begin{equation}
  \operatorname{nMAE}_h
  =
  \frac{1}{|\Omega_h|}
  \sum_{(i,j)\in\Omega_h}
  \left|
    \frac{\widehat x_{i,j,h}-\mu_j}{\sigma_j}
    -
    \frac{x_{i,j,h}-\mu_j}{\sigma_j}
  \right|.
  \label{eq:appendix_lab_nmae}
\end{equation}
Laboratory-availability AUPRC and Brier score the observation mask rather than
an unmeasured numeric value.

The bounded candidate-action probe reports tie-aware top-$k$ recovery of the
recorded action. Supplementary diagnostics additionally report risk-rule and
dose-token agreement where those fields are defined. These quantities measure
behavioral agreement within the supplied candidate set. Let $\mathcal I_{\mathrm{cov}}$ contain cases for which the
recorded action is represented in the bounded candidate set. If $a_i^\star$ is
the recorded action, $s_i(a)$ is the candidate burden score, and $s_{i,(k)}$
is the $k$th-smallest score for case $i$, then
\begin{equation}
  \operatorname{Agree@}k
  =
  \frac{1}{|\mathcal I_{\mathrm{cov}}|}
  \sum_{i\in\mathcal I_{\mathrm{cov}}}
  \mathbf 1\!\left[s_i(a_i^\star)\le s_{i,(k)}\right].
  \label{eq:appendix_action_topk}
\end{equation}
The non-strict inequality includes all candidates tied at the $k$th-lowest
score.
Metrics from different target types are never summed or averaged into a single
system score because their semantics and scales differ.

\subsection{Factual Readouts And Prediction Controls}
\label{app:outcome_configurations}
\label{app:outcome_sanity}

\paragraph{Clinical Outcomes, Readout Design, And Probability Accuracy.}
Tables~\ref{tab:core7_per_target_metrics}--\ref{tab:mimiciv_common_event7}
provide the per-target results underlying the aggregate clinical comparison in
Table~\ref{tab:targetbenchmark} and Figure~\ref{fig:core7_event7_radar}. Each
target is scored on its observed test support, and the macro metrics are
unweighted averages across the seven targets in the corresponding panel. The
panels retain their dataset-specific target horizons, so results are summarized
within each panel.

\begin{table*}[htbp]
  \centering
  \footnotesize
  \caption{Per-target MIMIC-III Core-7 test results at the native target horizons. Macro scores are unweighted averages across the seven targets; the AUROC rows correspond to Figure~\ref{fig:core7_event7_radar}(a).}
  \label{tab:core7_per_target_metrics}
  \vspace{.5em}
  \resizebox{\textwidth}{!}{\begin{tabular}{lrrrrrrrr}
\toprule
\multicolumn{9}{l}{\textbf{AUPRC $\uparrow$}} \\
Method & \shortstack{Hypercapnic\\acidemia} & \shortstack{Severe\\acidemia} & \shortstack{Creatinine\\AKI} & \shortstack{Urine-output\\AKI} & \shortstack{GCS\\decrease} & \shortstack{Incident\\CAM+} & \shortstack{24-hour\\mortality} & Macro AUPRC \\
\midrule
medDreamer & 0.0581 & 0.0262 & 0.0226 & 0.2236 & 0.0102 & 0.0401 & 0.0244 & 0.0579 \\
SMB-Structure & 0.0637 & 0.0312 & 0.0155 & 0.2484 & 0.0115 & 0.0597 & 0.0170 & 0.0638 \\
SepsisAgent & 0.0660 & 0.0404 & 0.0270 & 0.2518 & 0.0198 & 0.1003 & 0.0354 & 0.0772 \\
ACWM & 0.0647 & 0.0389 & 0.0227 & 0.2637 & 0.0194 & 0.0982 & 0.0261 & 0.0762 \\
PatientWM (Ours) & \textbf{0.2556} & \textbf{0.1747} & \textbf{0.1192} & \textbf{0.5179} & \textbf{0.0386} & \textbf{0.1092} & \textbf{0.3440} & \textbf{0.2227} \\
\midrule
\multicolumn{9}{l}{\textbf{AUROC $\uparrow$}} \\
Method & \shortstack{Hypercapnic\\acidemia} & \shortstack{Severe\\acidemia} & \shortstack{Creatinine\\AKI} & \shortstack{Urine-output\\AKI} & \shortstack{GCS\\decrease} & \shortstack{Incident\\CAM+} & \shortstack{24-hour\\mortality} & Macro AUROC \\
\midrule
medDreamer & 0.5012 & 0.4821 & 0.6079 & 0.5859 & 0.3527 & 0.4419 & 0.6192 & 0.5130 \\
SMB-Structure & 0.5405 & 0.5408 & 0.5124 & 0.6359 & 0.4710 & 0.5500 & 0.5380 & 0.5412 \\
SepsisAgent & 0.5790 & 0.6139 & 0.6632 & 0.6462 & 0.5671 & 0.6166 & 0.6479 & 0.6191 \\
ACWM & 0.5757 & 0.6136 & 0.6431 & 0.6452 & 0.4620 & 0.6251 & 0.5971 & 0.5945 \\
PatientWM (Ours) & \textbf{0.8143} & \textbf{0.8095} & \textbf{0.8278} & \textbf{0.8204} & \textbf{0.7239} & \textbf{0.7006} & \textbf{0.8850} & \textbf{0.7974} \\
\midrule
\multicolumn{9}{l}{\textbf{BRIER $\downarrow$}} \\
Method & \shortstack{Hypercapnic\\acidemia} & \shortstack{Severe\\acidemia} & \shortstack{Creatinine\\AKI} & \shortstack{Urine-output\\AKI} & \shortstack{GCS\\decrease} & \shortstack{Incident\\CAM+} & \shortstack{24-hour\\mortality} & Macro BRIER \\
\midrule
medDreamer & 0.2910 & 0.2654 & 0.2169 & 0.2753 & 0.0621 & 0.1851 & 0.2387 & 0.2192 \\
SMB-Structure & 0.2571 & 0.2306 & 0.2334 & 0.2267 & 0.0809 & 0.2863 & 0.2005 & 0.2165 \\
SepsisAgent & 0.2679 & 0.2379 & 0.2143 & 0.2555 & 0.0848 & 0.2277 & 0.2390 & 0.2181 \\
ACWM & 0.2828 & 0.2428 & 0.1940 & 0.2767 & 0.0958 & 0.2166 & 0.2559 & 0.2235 \\
PatientWM (Ours) & \textbf{0.1062} & \textbf{0.0668} & \textbf{0.0387} & \textbf{0.1711} & \textbf{0.0120} & \textbf{0.0741} & \textbf{0.0641} & \textbf{0.0761} \\
\bottomrule
\end{tabular}
}
\end{table*}

\begin{table*}[htbp]
  \centering
  \footnotesize
  \caption{Per-target MIMIC-III Common Event-7 test results. Macro scores are unweighted averages across the seven targets; the AUROC rows correspond to Figure~\ref{fig:core7_event7_radar}(b).}
  \label{tab:mimiciii_common_event7}
  \vspace{.5em}
  \resizebox{\textwidth}{!}{\begin{tabular}{lrrrrrrrr}
\toprule
Method & \shortstack{Any acute\\deterioration} & \shortstack{New invasive\\ventilation} & \shortstack{New\\pressor} & \shortstack{Circulation\\pressor} & \shortstack{Respiratory\\IMV} & \shortstack{Sepsis-3\\onset} & \shortstack{Dynamic ICU\\mortality} & Macro \\
\midrule
\multicolumn{9}{l}{\textbf{MIMIC-III Common Event-7 Test Panel: AUPRC $\uparrow$}} \\
medDreamer & 0.0423 & 0.0335 & 0.0185 & 0.0234 & 0.0276 & 0.0317 & 0.0178 & 0.0278 \\
SMB-Structure & \textbf{0.0733} & \textbf{0.0725} & 0.0302 & 0.0495 & \textbf{0.0518} & 0.0694 & 0.0187 & 0.0522 \\
SepsisAgent & 0.0656 & 0.0530 & \textbf{0.0477} & 0.0507 & 0.0416 & \textbf{0.0786} & 0.0250 & 0.0517 \\
ACWM & 0.0517 & 0.0324 & 0.0404 & 0.0501 & 0.0437 & 0.0668 & 0.0256 & 0.0444 \\
PatientWM (Ours) & 0.0691 & 0.0620 & 0.0409 & \textbf{0.0718} & 0.0483 & 0.0758 & \textbf{0.0505} & \textbf{0.0598} \\
\midrule
\multicolumn{9}{l}{\textbf{MIMIC-III Common Event-7 Test Panel: AUROC $\uparrow$}} \\
medDreamer & 0.4873 & 0.4981 & 0.4679 & 0.5058 & 0.5029 & 0.5276 & 0.5449 & 0.5049 \\
SMB-Structure & 0.5431 & 0.5707 & 0.5910 & 0.6387 & 0.5849 & \textbf{0.7157} & 0.5557 & 0.6000 \\
SepsisAgent & \textbf{0.6352} & 0.6554 & 0.6841 & 0.7040 & 0.6469 & 0.7135 & 0.6028 & 0.6631 \\
ACWM & 0.4869 & 0.4889 & 0.5822 & 0.6682 & 0.6001 & 0.6802 & 0.6090 & 0.5879 \\
PatientWM (Ours) & 0.6340 & \textbf{0.6984} & \textbf{0.6971} & \textbf{0.7493} & \textbf{0.6597} & 0.7044 & \textbf{0.6428} & \textbf{0.6837} \\
\midrule
\multicolumn{9}{l}{\textbf{MIMIC-III Common Event-7 Test Panel: Brier $\downarrow$}} \\
medDreamer & 0.2721 & 0.2604 & 0.3267 & 0.2306 & \textbf{0.2329} & 0.2884 & 0.2304 & 0.2631 \\
SMB-Structure & 0.2303 & 0.2576 & 0.2068 & 0.1818 & 0.3104 & \textbf{0.1280} & 0.2797 & 0.2278 \\
SepsisAgent & 0.2455 & 0.3527 & \textbf{0.1867} & 0.2775 & 0.2492 & 0.2184 & 0.1901 & 0.2457 \\
ACWM & 0.2483 & 0.2612 & 0.3150 & 0.1509 & 0.2351 & 0.2689 & 0.2313 & 0.2444 \\
PatientWM (Ours) & \textbf{0.2069} & \textbf{0.2197} & 0.1937 & \textbf{0.1369} & 0.2351 & 0.2151 & \textbf{0.1673} & \textbf{0.1964} \\
\bottomrule
\end{tabular}
}
\end{table*}

\begin{table*}[htbp]
  \centering
  \footnotesize
  \caption{Per-target MIMIC-IV Common Event-7 test results. Macro scores are unweighted averages across the seven targets; the AUROC rows correspond to Figure~\ref{fig:core7_event7_radar}(c).}
  \label{tab:mimiciv_common_event7}
  \vspace{.5em}
  \resizebox{\textwidth}{!}{\begin{tabular}{lrrrrrrrr}
\toprule
Method & \shortstack{Any acute\\deterioration} & \shortstack{New invasive\\ventilation} & \shortstack{New\\pressor} & \shortstack{Circulation\\pressor} & \shortstack{Respiratory\\IMV} & \shortstack{Sepsis-3\\onset} & \shortstack{Dynamic ICU\\mortality} & Macro \\
\midrule
\multicolumn{9}{l}{\textbf{MIMIC-IV Common Event-7 Test Panel: AUPRC $\uparrow$}} \\
medDreamer & 0.0653 & 0.0428 & 0.0253 & 0.0257 & 0.0346 & 0.0272 & 0.0178 & 0.0341 \\
SMB-Structure & 0.0901 & 0.0589 & 0.0548 & 0.0514 & 0.0528 & 0.0494 & 0.0657 & 0.0604 \\
SepsisAgent & 0.2733 & 0.1193 & 0.2093 & \textbf{0.2044} & 0.1228 & 0.0819 & 0.2087 & 0.1742 \\
ACWM & 0.2725 & 0.1202 & \textbf{0.2107} & 0.2042 & 0.1259 & 0.0827 & 0.2214 & 0.1768 \\
PatientWM (Ours) & \textbf{0.2816} & \textbf{0.1253} & 0.2066 & 0.1970 & \textbf{0.1265} & \textbf{0.0843} & \textbf{0.2278} & \textbf{0.1784} \\
\midrule
\multicolumn{9}{l}{\textbf{MIMIC-IV Common Event-7 Test Panel: AUROC $\uparrow$}} \\
medDreamer & 0.5347 & 0.5111 & 0.4734 & 0.4891 & 0.5293 & 0.5094 & 0.4877 & 0.5050 \\
SMB-Structure & 0.5994 & 0.5933 & 0.6661 & 0.6635 & 0.6044 & 0.6460 & 0.6898 & 0.6375 \\
SepsisAgent & 0.8105 & 0.7341 & 0.8416 & 0.8389 & 0.7535 & \textbf{0.7478} & 0.8531 & 0.7971 \\
ACWM & 0.8089 & 0.7339 & 0.8404 & 0.8378 & 0.7556 & 0.7446 & 0.8504 & 0.7959 \\
PatientWM (Ours) & \textbf{0.8124} & \textbf{0.7375} & \textbf{0.8428} & \textbf{0.8394} & \textbf{0.7563} & 0.7477 & \textbf{0.8552} & \textbf{0.7988} \\
\midrule
\multicolumn{9}{l}{\textbf{MIMIC-IV Common Event-7 Test Panel: Brier $\downarrow$}} \\
medDreamer & 0.2817 & 0.2631 & 0.2595 & 0.2519 & 0.2555 & 0.2181 & 0.2528 & 0.2547 \\
SMB-Structure & 0.2889 & 0.2566 & 0.2182 & 0.2314 & 0.3242 & 0.2821 & 0.1987 & 0.2572 \\
SepsisAgent & 0.1933 & 0.2746 & 0.1404 & \textbf{0.1408} & 0.2360 & 0.2004 & 0.1266 & 0.1875 \\
ACWM & 0.1976 & 0.2632 & 0.1436 & 0.1465 & 0.2257 & \textbf{0.1944} & \textbf{0.1212} & 0.1846 \\
PatientWM (Ours) & \textbf{0.1699} & \textbf{0.2216} & \textbf{0.1387} & 0.1414 & \textbf{0.1845} & 0.2007 & 0.1358 & \textbf{0.1704} \\
\bottomrule
\end{tabular}
}
\end{table*}

\paragraph{External Point Estimates.}
Figure~\ref{fig:external_transfer_main} in the main text summarizes the external
AUROC comparison. Table~\ref{tab:external_transfer_main} reports the exact
test AUPRC, AUROC, and Brier point estimates. PatientWM leads the listed
adapted clinical world-model baselines on all three datasets.

\begin{table*}[htbp]
  \centering
  \footnotesize
  \caption{External test evaluation of PatientWM and available clinical-model baselines. All rows use the same per-target observed test support within each dataset. Bold marks the best value within each dataset and metric.}
  \label{tab:external_transfer_main}
  \vspace{.5em}
  \begin{tabular}{llrrr}
\toprule
Dataset & Method & Macro AUPRC $\uparrow$ & Macro AUROC $\uparrow$ & Brier $\downarrow$ \\
\midrule
ALOTT & medDreamer & 0.0675 & 0.5659 & 0.0474 \\
 & SMB-Structure & 0.0651 & 0.5672 & 0.0525 \\
 & SepsisAgent & 0.2007 & 0.7864 & 0.0451 \\
 & ACWM & 0.1838 & 0.7757 & 0.0451 \\
 & PatientWM (Ours) & \textbf{0.2268} & \textbf{0.7915} & \textbf{0.0437} \\
\midrule
VitalDB & medDreamer & 0.0538 & 0.5082 & 0.0441 \\
 & SMB-Structure & 0.0540 & 0.5160 & 0.0545 \\
 & SepsisAgent & 0.0859 & 0.6114 & 0.0436 \\
 & ACWM & 0.0868 & 0.6264 & 0.0435 \\
 & PatientWM (Ours) & \textbf{0.1000} & \textbf{0.6725} & \textbf{0.0426} \\
\midrule
MC-MED & medDreamer & 0.2049 & 0.5039 & 0.1485 \\
 & SMB-Structure & 0.2493 & 0.5726 & 0.1963 \\
 & SepsisAgent & 0.3239 & 0.6600 & 0.1391 \\
 & ACWM & 0.3232 & 0.6565 & 0.1384 \\
 & PatientWM (Ours) & \textbf{0.3423} & \textbf{0.6764} & \textbf{0.1352} \\
\bottomrule
\end{tabular}

\end{table*}

\subsection{Rollout And Action-Planning Details}
\label{app:rollout_details}
\label{app:scorer_details}

\paragraph{Multi-Method Recursive Rollout.}
Figure~\ref{fig:mimiciv_strict_rollout} in the main text summarizes the
multi-method rollout comparison. Table~\ref{tab:mimiciv_strict_rollout} reports
the exact Event-7 and laboratory values below. All methods are evaluated on the
same trajectories, horizons, recorded actions, targets, and evaluator.

\begin{table*}[htbp]
  \centering
  \footnotesize
  \caption{MIMIC-IV multi-step rollout on the same test trajectories and targets. We report recursive Event-7 macro-AUPRC and observed-laboratory nMAE from the typed laboratory decoder. Bold marks the best value in each column.}
  \label{tab:mimiciv_strict_rollout}
  \vspace{.5em}
  \resizebox{0.98\textwidth}{!}{\begin{tabular}{lrrrrrrrr}
\toprule
& \multicolumn{4}{c}{Event-7 macro-AUPRC $\uparrow$}
& \multicolumn{4}{c}{Observed-laboratory nMAE $\downarrow$} \\
\cmidrule(lr){2-5}\cmidrule(lr){6-9}
Method & 6 h & 24 h & 48 h & 96 h & 6 h & 24 h & 48 h & 96 h \\
\midrule
medDreamer
  & 0.0474 & 0.0733 & 0.1038 & 0.1052
  & 0.4243 & 0.4527 & 0.5093 & 0.5728 \\
SMB-Structure
  & 0.1141 & 0.0546 & 0.0493 & 0.0425
  & 0.6528 & 0.7082 & 0.8018 & 0.9533 \\
SepsisAgent
  & 0.1637 & 0.0844 & 0.0927 & 0.0939
  & 0.4191 & 0.4520 & 0.5085 & 0.5690 \\
ACWM
  & 0.1378 & 0.0665 & 0.0761 & 0.0714
  & 0.4240 & 0.4519 & 0.5084 & 0.5729 \\
PatientWM (Ours)
  & \textbf{0.1755} & \textbf{0.1005} & \textbf{0.1090} & \textbf{0.1308}
  & \textbf{0.4093} & \textbf{0.4161} & \textbf{0.4556} & \textbf{0.4958} \\
\bottomrule
\end{tabular}
}
\end{table*}

\paragraph{Action-Planning Evaluation.}

Table~\ref{tab:prediction_only_action_scoring} summarizes the main-text result.
The action-planning evaluation includes 15,951 test trajectories for each method.
Its candidate provenance hash is identical across methods. Candidate actions
come only from a prespecified empty update or the same patient's earlier
recorded history. The current action, actions from other patients, and future
outcomes are not used to create or score candidates. All methods use the
source-boundary \texttt{boundary\_known} action view. The primary denominator is
13,143 test decisions with non-empty recorded treatment updates. Item-level
candidate coverage is 0.5573 (95\% patient-cluster interval 0.5458--0.5672);
coverage across all decisions, including empty updates, is 0.6352. Ties are retained. The event
score averages predicted sepsis-onset and cardiac-arrest probabilities. The
laboratory score is the predicted absolute standardized value, weighted by
predicted measurement probability. The combined score first scales both
burdens within each candidate set and then adds a fixed uncertainty penalty
shared by all methods. The main comparison uses shared host inputs, heads, and
evaluator for all methods. SMB-Structure passes its frozen predictive state
through an action-aware transition before scoring each candidate. Under the
fixed score, PatientWM obtains the highest covered top-1 and top-2 agreement.

\paragraph{Surrogate Boundary.}
This diagnostic follows a common limitation of retrospective treatment-planning
evaluations: recorded clinician regimens, expert preference, guideline
adherence, and off-policy value estimates can support behavioral plausibility,
but they do not directly reveal the true outcome of an unexecuted action for
the same patient. The observed record contains the outcome under historical
care, while alternative candidate outcomes remain unobserved. We therefore
interpret recorded-action agreement as a clinician-behavior surrogate, not as
evidence of treatment optimality, counterfactual correctness, or patient-level
action quality. Action-conditioned systems such as SepsisAgent and CLARITY
\citep{wu2026sepsisagent,ding2026clarity} can query alternative actions and
report policy or regimen surrogates, but without additional causal
identification, support diagnostics, or randomized-like validation,
model-predicted risk or agreement with historical or expert choices does not by
itself establish treatment benefit.

\subsection{Frozen-State Reuse}
\label{app:frozen_state_reuse}
Table~\ref{tab:frozen_state_transfer} reports the exact values summarized in
Figure~\ref{fig:frozen_state_transfer}. The frozen PatientWM representation is
trained without clinical-event label loss and evaluated with a newly trained
Common Event-7 readout. All rows use the same seven targets, 14,044 test
examples, per-target observed and positive counts, and evaluator. The full
PatientWM row provides an in-family reference for the cost of removing event
supervision during source training; the remaining rows reproduce the baseline
results from Table~\ref{tab:targetbenchmark}.

\begin{table*}[htbp]
  \centering
  \footnotesize
  \caption{MIMIC-III Common Event-7 frozen-state transfer on the
  Table~\ref{tab:targetbenchmark} test support. Bold marks the best result in
  each column.}
  \label{tab:frozen_state_transfer}
  \vspace{.5em}
  \resizebox{0.98\textwidth}{!}{\begin{tabular}{lrrr}
\toprule
Method & Macro AUPRC $\uparrow$ & Macro AUROC $\uparrow$ & Macro Brier $\downarrow$ \\
\midrule
medDreamer & 0.0278 & 0.5049 & 0.2631 \\
SMB-Structure & {0.0522} & 0.6000 & {0.2278} \\
SepsisAgent & 0.0517 & {0.6631} & 0.2457 \\
ACWM & 0.0444 & 0.5879 & 0.2444 \\
PatientWM (Ours) & \textbf{0.0598} & \textbf{0.6837} & \textbf{0.1964} \\
PatientWM (Frozen transfer) & 0.0500 & 0.6122 & 0.2401 \\
\bottomrule
\end{tabular}
}
\end{table*}

\subsection{Label-Efficient Cross-Dataset Adaptation}
\label{app:low_data_adaptation}

The cross-dataset experiment transfers the Common Event-7 prediction
interface from MIMIC-III to MIMIC-IV while keeping each source predictor
frozen. For every method, the only target-domain component learned is the same
residual logit adapter with 56 trainable parameters. The adapter is trained on
328 examples, corresponding to 1\% of a fixed 32,768-example MIMIC-IV training
support. All methods use the same target examples, seven outcome
definitions, and evaluator. This protocol isolates label-efficient
cross-cohort reuse:
the source representation and prediction interface remain fixed, while the
small target-domain adapter accounts for cohort-level shifts in the output
space. The aggregate MIMIC-IV test results are reported in
Table~\ref{tab:low_data_generalization}.

\section[Reproducibility And Artifacts]{Reproducibility And Artifacts}

We report the data splits and model parameters used for the principal
experiments. The MIMIC-III transition model uses all 32 training shards with
equal contribution and is evaluated on all 32 test shards. The test split
contains 3,827 subjects, 5,337 ICU episodes, and 14,317 transition
trajectories; paired waveform support is 2,291/1,911/1,401/877 targets at
6/24/48/96 hours. The model uses micro-batches of eight with four-step gradient
accumulation, AdamW learning rate $3\times10^{-4}$, weight decay 0.01, and a
closed-loop objective that advances up to 16 six-hour windows from one observed
initial state.

The MIMIC-IV transition model for recursive rollout and action-planning
evaluation uses 15,995 validation trajectories and 15,951
common test trajectories across all 32 shards. It uses
a two-layer width-128 GRU, dropout 0.2, context length 16, ordered state--action
history, concatenated transition fusion, and gated residual latent update. It
trains with batch size eight, AdamW learning rate $3\times10^{-4}$, weight
decay 0.01, and a 16-window closed-loop term weighted by 0.50.

For the MIMIC-IV laboratory results in Table~\ref{tab:mimiciv_matched_test}, the
typed laboratory decoder trains with the transition
and GapShrink prior frozen. It has 36,483 trainable parameters, a
32-dimensional item query, dual item--cross-laboratory attention, and
source-only trend, history, count, window-summary, and laboratory-ledger
features. The decoder is selected by validation all-target laboratory nMAE.
Exact common test support is 246,487/149,654/105,953/58,858 observed
item-targets at 6/24/48/96 hours.

For the MIMIC-III Core-7 results in Table~\ref{tab:targetbenchmark}, the
clinical-outcome readout uses 2,283 source features and appends 228
source-history features for 2,511 inputs. The appended source-history features
are available for 106,753 of 107,859 training windows and 14,033 of 14,122 test
windows; the remaining rows retain the availability indicator off. Each
outcome uses 120 histogram-gradient-boosting iterations,
31 leaves, learning rate 0.05, minimum leaf size 20, $L_2=1$, and balanced
class weights.

For the external-dataset evaluations in Figure~\ref{fig:external_transfer_main},
the models use context length eight, shared host width 192,
batch size 64, and AdamW learning rate $2\times10^{-4}$. PatientWM uses three
layers and six attention heads. SepsisAgent, medDreamer, and ACWM receive the
same host state and known-empty external-action port, while SMB-Structure maps
the same host representation through its frozen predictive core. Every learned
row shares a balanced binary outcome loss plus 0.10-weighted next-EHR and
next-waveform SmoothL1 losses where adjacent targets exist. Dataset-specific
patient-group training, validation, and test splits are used for all external
evaluations.

\end{document}